\documentclass{article} 
\usepackage[preprint]{iclr2027_conference}
\usepackage{times}

\usepackage{amsmath,amsfonts,bm}

\def\eqref#1{equation~\ref{#1}}

\def\1{\bm{1}}

\DeclareMathAlphabet{\mathsfit}{\encodingdefault}{\sfdefault}{m}{sl}
\SetMathAlphabet{\mathsfit}{bold}{\encodingdefault}{\sfdefault}{bx}{n}

\usepackage{amsmath}  
\usepackage{amssymb}  
\usepackage{algorithmicx,algorithm}
\usepackage{amsmath}
\usepackage{xspace}
\usepackage[utf8]{inputenc} 
\usepackage[T1]{fontenc}    
\usepackage{hyperref}
\usepackage{url}
\usepackage{booktabs}       
\usepackage{amsfonts}       
\usepackage{nicefrac}       
\usepackage{footmisc}
\usepackage[nopatch=footnote]{microtype}
\usepackage{xcolor}         
\usepackage{multirow}
\usepackage{array}       
\usepackage{tabularx}    
\usepackage{algpseudocode}                  
\usepackage{setspace}                       
\usepackage{lipsum} %
\usepackage{subcaption} 
\usepackage{enumitem}
\usepackage{caption} 
\usepackage{colortbl} 
\usepackage{bbm}
\usepackage{algorithm}
\usepackage{graphicx}
\usepackage{xcolor}
\usepackage{relsize}
\usepackage{etoolbox}
\usepackage{wrapfig} 
\usepackage{wrapfig}
\usepackage{graphicx}
\usepackage{caption}
\usepackage[dvipsnames]{xcolor}
\usepackage{pifont}
\definecolor{gyk}{HTML}{7B2CBF}

\newcommand{\name}{CoVA}
\newcommand{\extyes}{\textcolor{green!60!black}{\ding{51}}}
\newcommand{\extno}{\textcolor{red}{\ding{55}}}
\newcommand{\pz}

\title{CoVA: Composing VAE Representations for Generative Learning}

\author{
  Zhaowei Chen$^{1}$\quad
  Yikai Ge$^{2}$\quad
  Linze Li$^{1*}$\quad
  Yao Tang$^{1}$\quad
  Xiang Li$^{2}$\\[0.04cm]
  $^{1}$Jiiov Technology,
  $^{2}$Nankai University
}

\renewcommand{\thefootnote}{\fnsymbol{footnote}}

\definecolor{softblue}{RGB}{225,235,247}
\definecolor{softyellow}{RGB}{249,241,205}
\definecolor{softorange}{RGB}{249,226,205}
\definecolor{softgreen}{RGB}{224,240,226}
\definecolor{softpurple}{RGB}{235,226,243}
\definecolor{softgray}{RGB}{235,235,235}

\begin{document}

\maketitle
\footnotetext[1]{Corresponding author.}
\renewcommand{\thefootnote}{\arabic{footnote}}
\setcounter{footnote}{0}

\begin{figure}[ht]
    \centering
    \includegraphics[width=0.9\linewidth]{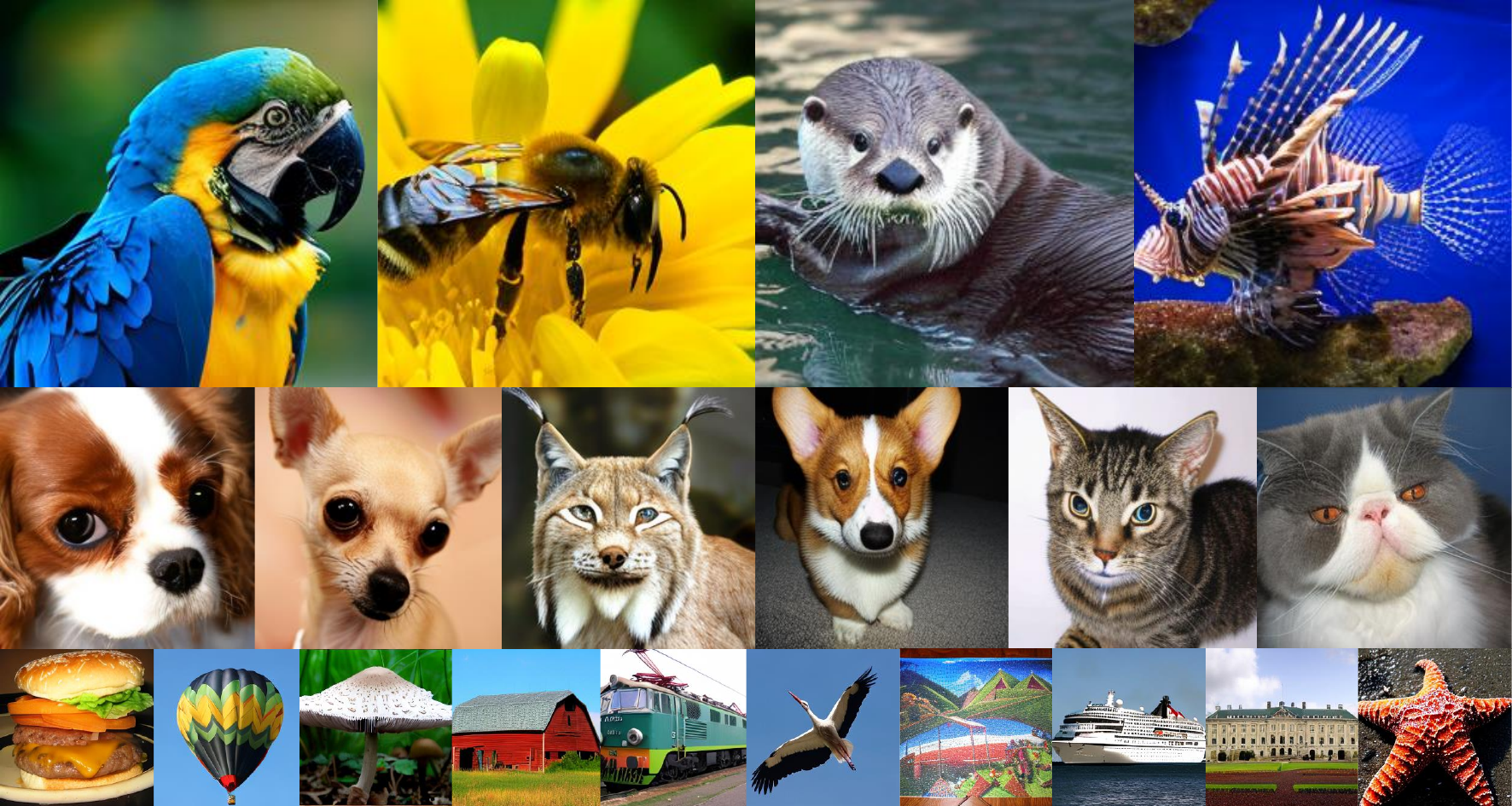}
    \vspace{-5pt}
    \caption{\textbf{High-fidelity image generation with CoVA.} Samples from SiT-XL/2 with CoVA on ImageNet at $256\times256$ resolution. CoVA produces images with fine-grained details, coherent structures, and strong semantic consistency.}
    \label{fig:abstract-fig}
    \vspace{-1em}
\end{figure}

\begin{abstract}
Representation alignment with pretrained visual encoders improves diffusion transformer training but requires auxiliary encoders and extra training overhead. Recent approaches instead exploit representations from the generative model's own variational autoencoder (VAE). However, reconstruction-oriented VAE features are not optimized for the semantic discrimination and spatial structure needed for generative learning. We introduce \textbf{CoVA} (\textbf{Co}mposed \textbf{VA}E Representations), a lightweight framework that composes VAE features online into stage-wise guidance for flow-matching training. A small latent composer is jointly trained with class supervision, masked latent reconstruction, and horizontal-flip consistency, yielding semantically discriminative, spatially structured dense features and sample-dependent class posteriors. CoVA aligns intermediate SiT features with the composed representations early in training, then removes the increasingly detrimental alignment and reuses detached composer posteriors for sample-dependent soft label conditioning. We further uncover a hidden confound in soft label conditioning: averaging class embeddings attenuates their magnitude, coupling semantic interpolation with conditioning strength. This attenuation alone can substantially improve FID while degrading Recall, revealing a non-semantic source of the apparent gain. CoVA corrects this confound by restoring the hard-label magnitude, isolating posterior-dependent directional semantics. CoVA adds only 8.5M training-only auxiliary parameters (1.3\% of SiT-XL/2) and introduces no inference-time overhead. On ImageNet 256$\times$256, CoVA reduces SiT-XL/2 FID from 17.2 to 7.4 (\textbf{57\%}) in 400K steps. With classifier-free guidance, it achieves an FID of \textbf{2.03} in only 400 epochs, versus 2.06 after 1,400 epochs for vanilla SiT-XL/2.
\end{abstract}

\begin{figure*}[t]
    \centering
    \includegraphics[width=1.0\linewidth]{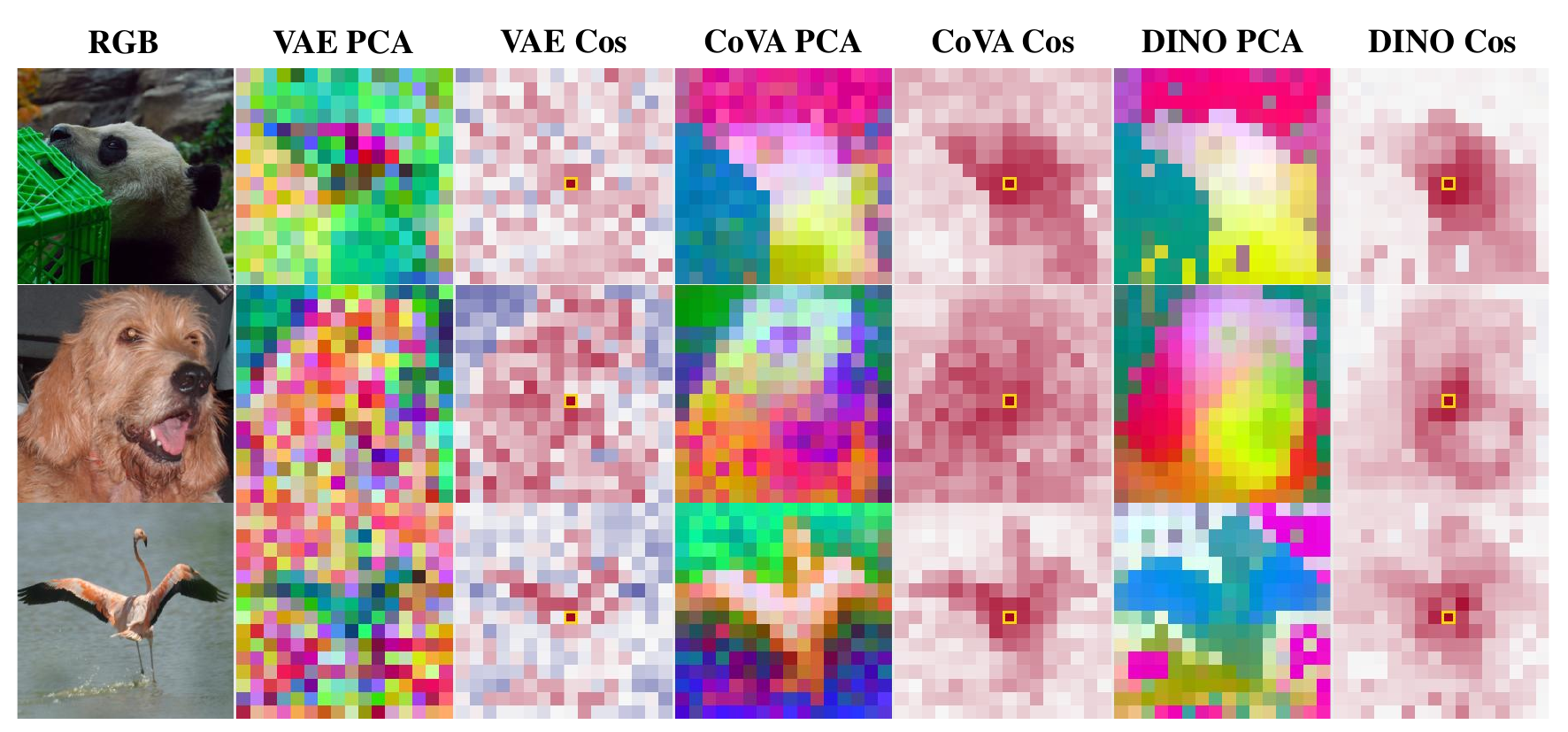}
    \caption{\textbf{CoVA reorganizes raw VAE features into more spatially coherent representations.}
    Each row shows one ImageNet image together with PCA visualizations and cosine-similarity maps of raw VAE latent features, CoVA-composed features, and DINOv2-B/14 features.
    For PCA visualization, patch features are L2-normalized and projected onto the first three principal components, which are mapped to RGB channels.
    Each cosine map shows the similarity between all patch tokens and the query token marked by the yellow box.
    Compared with raw VAE features, CoVA produces substantially more coherent and object-aligned spatial structure.
    Quantitatively, Spatial LDS($\uparrow$) improves from $0.28$ for raw VAE features to $0.41$ with CoVA, comparable to the $0.40$ achieved by DINOv2.}
    \label{fig:PCA}
    \vspace{-10pt}
\end{figure*}

\section{Introduction}
Recent advances in diffusion and flow-matching models have substantially reshaped modern image generation, delivering strong gains in visual quality, scalability, and training efficiency. From the original DDPM formulation \cite{ddpm} and accelerated sampling methods \cite{ddim} to flow-matching objectives \cite{fm} and scalable transformer architectures such as DiT \cite{dit} and SiT \cite{sit}, these developments have steadily pushed the frontier of generative modeling. Yet recent evidence suggests that generation quality depends not only on the learning objective, but also on the structure of intermediate representations formed during optimization. Representation alignment methods such as REPA \cite{repa} and its follow-up analyses \cite{irepa} have shown that explicitly regularizing intermediate generative features can accelerate training and improve generation quality. A common strategy is to align these features with representations extracted from pretrained visual encoders such as DINO \cite{dino} or CLIP \cite{clip}. While effective, this approach requires an additional visual backbone throughout training, introducing nontrivial computational and memory overhead.

A natural alternative is to reuse representations available within the latent generative pipeline. Recent work has shown that VAE-derived features can provide effective intermediate supervision without relying on an external visual encoder \cite{repa-e,sra2}. However, VAE representations are learned primarily for reconstruction rather than semantic organization \cite{rae}, and are therefore not necessarily optimal supervision targets for a generative transformer. This motivates a fundamental question: \emph{can reconstruction-oriented VAE features be lightly reorganized into representations that are better suited for generative learning?} To address this, we introduce \textbf{CoVA} (\textbf{Co}mposed \textbf{VA}E Representations), a lightweight framework that learns to compose VAE representations online during flow-matching training. CoVA augments the VAE features with a latent composer trained using complementary semantic and structural objectives, including class supervision, masked latent reconstruction \cite{mae}, and horizontal-flip consistency. As illustrated in Fig.~\ref{fig:PCA}, this lightweight composition transforms the weakly organized spatial structure of raw VAE latents into coherent, object-aligned representations. Motivated by the finding of iREPA~\cite{irepa} that spatial structure is critical to effective representation alignment, we use these composed dense features as supervisory targets for an intermediate SiT layer through a lightweight projector. The composer is trained online alongside the generative model using signals available in the training pipeline, avoiding dependence on a pretrained DINO-style or CLIP-style teacher.

Our experiments further reveal that even improved representation supervision is not uniformly beneficial throughout optimization. Composed VAE features provide clear gains during the earlier stages of training, whereas prolonged alignment eventually degrades generation performance, consistent with recent observations \cite{haste} on REPA. This suggests that representation guidance is better viewed as a stage-dependent signal than as a persistent constraint. It also raises a natural follow-up question: \emph{once explicit alignment ceases to help, can the semantic information learned by the composer still benefit the generator without constraining its internal feature space?} CoVA addresses this by shifting the role of the composer from representation supervision to conditioning. At later stages, CoVA removes the intermediate alignment objective and reuses the composer's detached class posterior for sample-dependent Soft Label Conditioning. The VAE-derived signal thus evolves from shaping internal features to modulating class conditioning as training progresses.

A closer analysis reveals a hidden confound in naive soft label conditioning. Forming a soft condition by averaging class embeddings changes two factors simultaneously: its semantic direction and its magnitude. The latter unintentionally alters the strength of class conditioning, making it difficult to distinguish gains from semantic interpolation from those caused by weaker or stronger conditioning. Empirically, naive soft conditioning lowers FID \cite{fid} and improves Precision, but also reduces Recall \cite{rec}, revealing a clear quality--diversity trade-off. Controlled ablations show that this trade-off is largely driven by the unintended magnitude change. CoVA therefore explicitly restores the condition magnitude, removing this confounding effect and allowing the posterior to contribute primarily through its sample-dependent semantic direction.

CoVA introduces only 8.5M (1.3\% of SiT-XL/2) training-only auxiliary parameters, including a 5.6M latent composer and a 2.9M alignment projector, incurs only 7.2\% additional training-time GFLOPs, and is discarded at inference. Across SiT model scales, CoVA consistently improves generation quality under matched training budgets. On ImageNet, it reduces SiT-XL/2 FID from 17.2 to 7.4 at 400K steps, while reaching 2.03 with classifier-free guidance in only 400 epochs. These results demonstrate that VAE representations, when appropriately composed and reused across training stages, can provide effective generative supervision without auxiliary visual encoders.

\noindent
{In summary, our contributions are as follows:}
\vspace{-5pt}
\begin{itemize}
\item We introduce a lightweight online latent composer that reorganizes
reconstruction-oriented VAE latents into semantically discriminative and
spatially structured dense representations, providing effective intermediate
supervision without an external visual encoder.

\item Building on the stage-dependent nature of representation alignment,
we show that VAE-derived information can play complementary roles across
optimization. CoVA uses composed dense features for early representation
shaping, then removes the feature constraint and later reuses the same
composer through its detached, sample-dependent class posterior.

\item We identify a hidden magnitude confound in soft label conditioning:
posterior averaging changes not only semantic direction but also condition
magnitude, creating an unintended conditioning-strength shortcut. By restoring
the hard-label magnitude, CoVA isolates the posterior-dependent directional
signal and substantially mitigates the resulting fidelity--coverage trade-off.

\vspace{-5pt}
\end{itemize}

\section{Related work}
\noindent\textbf{Diffusion Transformers and Flow-Matching Models.}
Diffusion models have evolved from pixel-space denoising formulations~\cite{adm,vdm,simple} to latent generative models~\cite{ldm}, substantially reducing the computational cost of high-resolution synthesis. The adoption of transformer backbones~\cite{attention,vit,u-vit,diffit,mdtv2,maskdit,sd-dit} has further improved scalability, leading to architectures such as DiT~\cite{dit}, SiT~\cite{sit}, PixArt-$\alpha$~\cite{pixartalpha}, Stable Diffusion 3~\cite{sd3}, and Lumina~\cite{lumina2}. In particular, DiT establishes transformer-based denoising in latent space, while SiT generalizes this design to interpolant-based and flow-based objectives. Our work builds on SiT and focuses not on modifying the generative objective or backbone, but on improving how useful representations and conditioning signals are formed during training.

\begin{figure}[t]
    \centering
    \includegraphics[width=1.0\linewidth]{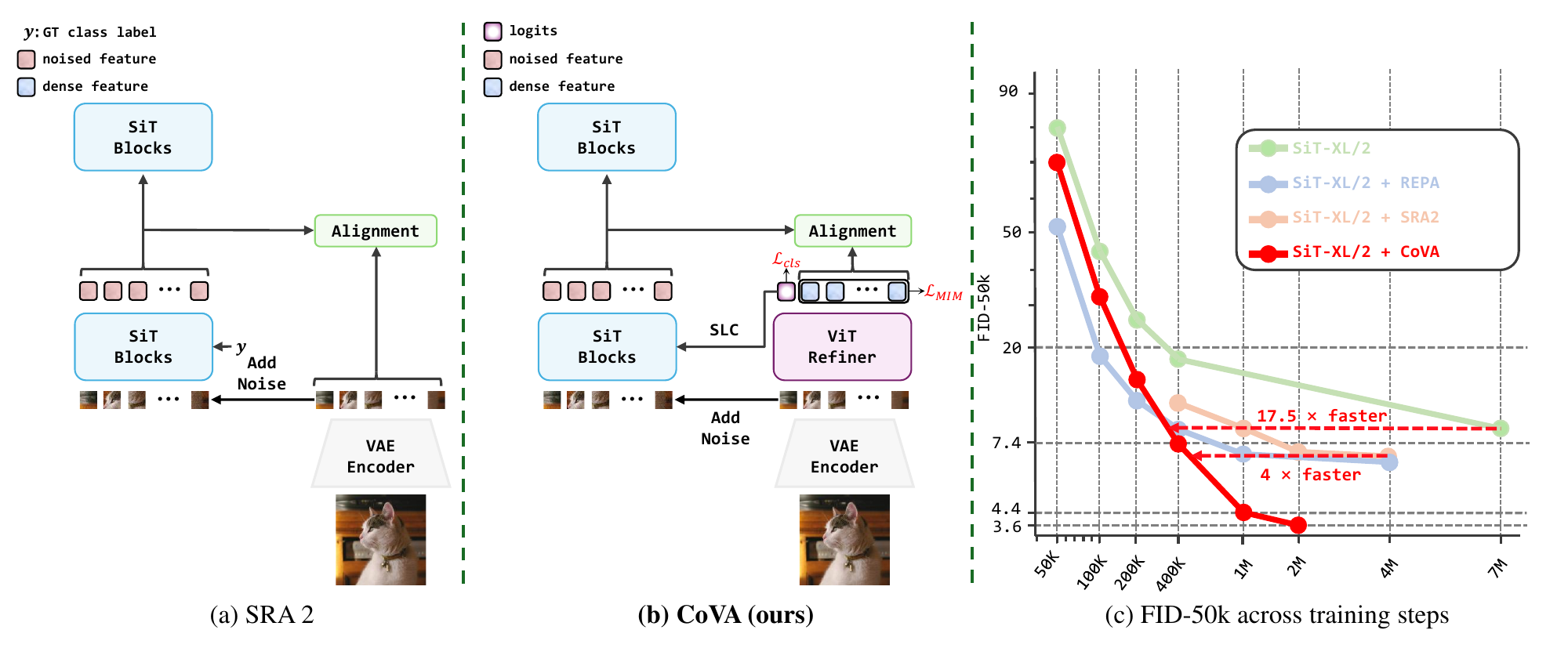}
    \caption{\textbf{Comparison between SRA2 and CoVA.} 
    (a) SRA2 aligns intermediate SiT features with VAE features through a
    lightweight projection layer and an auxiliary feature-alignment loss.
    (b) CoVA first composes clean VAE latents with an online composer trained
    by semantic and spatial objectives, and then reuses its dense features for
    early representation alignment and its detached class posterior for
    later soft conditioning.
    (c) CoVA's stage-adaptive transition from representation alignment to soft
    conditioning enables faster training convergence and improved generation
    quality.
    }
    \label{fig:main}
    \vspace{-10pt}
\end{figure}

\noindent\textbf{Intermediate Representation Supervision for Diffusion Training.}
Recent work has shown that explicitly shaping intermediate representations can substantially accelerate diffusion-transformer training. REPA~\cite{repa} aligns intermediate diffusion features with representations from pretrained visual encoders, while subsequent studies further investigate which properties of the target representation are most beneficial for generation~\cite{irepa,reg}. In particular, iREPA highlights the importance of spatial structure in the alignment target rather than relying solely on global semantic quality~\cite{irepa}. Although effective, these methods require pretrained representation encoders throughout training. Other approaches instead derive supervision internally. SRA~\cite{sra}, DoD~\cite{dod}, and SSD~\cite{ssd}, for example, construct guidance from representations produced by the generative model or related teacher branches, reducing reliance on external foundation encoders. HASTE~\cite{haste} further shows that representation alignment is strongly stage dependent: REPA provides substantial early benefits but can plateau or become detrimental when maintained later in training. CoVA shares the motivation of lightweight and stage-aware representation learning, but differs in both the source and the use of supervision: it first learns a task-adapted representation from VAE features and subsequently reuses the learned semantic information through class conditioning rather than maintaining feature alignment.

\noindent\textbf{VAE-Derived Representations for Diffusion Training.}
Variational autoencoders are fundamental components of latent diffusion systems, providing compact image representations that can be efficiently modeled by generative networks~\cite{ldm,vae}. Existing studies have explored VAE-based improvements from two complementary perspectives. Latent compression methods~\cite{dc-ae,dc-ae2} reduce the spatial or channel redundancy of latent tokens to improve computational efficiency, while representation-oriented methods~\cite{vavae,xu2025exploring,repa-e} aim to make the latent space more suitable for generation. For example, REPA-E~\cite{repa-e} jointly optimizes the VAE tokenizer and the latent diffusion model, whereas VAVAE~\cite{vavae} investigates the tension between reconstruction-oriented representations and generation-oriented objectives. SRA 2~\cite{sra2} demonstrates that VAE features can also provide effective visual guidance for diffusion transformers through a lightweight projection-based alignment objective, without introducing an additional large visual encoder or a second diffusion model. CoVA is most closely related to this direction, but differs in two important respects. First, rather than aligning the generator directly to raw reconstruction-oriented VAE features, CoVA learns a lightweight online composer that reorganizes them using semantic and structural objectives before they are used as supervision. Second, CoVA does not retain VAE-based feature alignment as a fixed training constraint: once alignment ceases to be beneficial, the composer's classifier posterior is reused for sample-dependent class conditioning.

\section{Method}

\subsection{Preliminaries}
\label{sec:3.1}

Our method is built upon Scalable Interpolant Transformers (SiT)~\cite{sit}, which provide a common formulation for flow-based and diffusion-based generation. SiT constructs a continuous transport path between a Gaussian prior and the data distribution, and learns the corresponding vector field with a velocity-matching objective.

Let $\boldsymbol{z}\sim p_{\mathrm{data}}$ denote a clean latent sample and let $\boldsymbol{\epsilon}\sim\mathcal{N}(\boldsymbol{0},\boldsymbol{I})$ be an independent Gaussian noise sample. We use the linear interpolation
\begin{equation}
\boldsymbol{y}_t=t\boldsymbol{z}+(1-t)\boldsymbol{\epsilon},
\qquad t\in[0,1],
\end{equation}
which connects Gaussian noise at $\boldsymbol{y}_0=\boldsymbol{\epsilon}$ to the clean latent at $\boldsymbol{y}_1=\boldsymbol{z}$. Its instantaneous target velocity is $\boldsymbol{u}_t=\boldsymbol{z}-\boldsymbol{\epsilon}$.

SiT parameterizes the velocity field with a neural network $\boldsymbol{v}_{\boldsymbol{\phi}}(\boldsymbol{y}_t,t,\boldsymbol{c})$, where $\boldsymbol{c}$ denotes the class condition. The base flow-matching objective is
\begin{equation}
\mathcal{L}_{\mathrm{FM}}(\boldsymbol{c})
=
\mathbb{E}_{t\sim\mathcal{U}[0,1],\boldsymbol{z},\boldsymbol{\epsilon}}
\left[
\frac{1}{n}
\left\|
\boldsymbol{v}_{\boldsymbol{\phi}}(\boldsymbol{y}_t,t,\boldsymbol{c})
-
\left(\boldsymbol{z}-\boldsymbol{\epsilon}\right)
\right\|_2^2
\right],
\end{equation}
where $n$ is the number of latent elements. Generation proceeds from noise toward the data endpoint using the learned field. Our method retains the original SiT formulation and instead changes how VAE-derived information is used during training: first as composed intermediate supervision, and later as posterior-based class conditioning.

\subsection{Composing VAE Representations}
\label{sec:composer}

Directly using VAE features as representation targets is efficient, but these features are learned primarily for reconstruction rather than semantic organization. CoVA therefore learns a lightweight online composer that reorganizes the clean VAE latent into a representation better suited for intermediate supervision.

Let
$\boldsymbol{Z}\in\mathbb{R}^{N\times d_z}$
denote the patchified clean VAE latent, where $N$ is the number of spatial latent patches and $d_z$ is the patch dimension.
A lightweight ViT-based composer~\cite{vit}
$F_{\theta}$ produces dense patch features
$\boldsymbol{H}\in\mathbb{R}^{N\times d_h}$
and class logits
$\boldsymbol{o}\in\mathbb{R}^{K}$:
\begin{equation}
(\boldsymbol{H},\boldsymbol{o})
=
F_{\theta}(\boldsymbol{Z}),
\qquad
\boldsymbol{H}
=
[\boldsymbol{h}_1,\ldots,\boldsymbol{h}_N],
\end{equation}
where $K$ denotes the number of classes.

The composer is trained online with complementary semantic and spatial objectives:
\begin{equation}
\mathcal{L}_{\mathrm{comp}}
=
\mathcal{L}_{\mathrm{cls}}
+
\lambda_{\mathrm{mim}}\mathcal{L}_{\mathrm{mim}}
+
\lambda_{\mathrm{flip}}\mathcal{L}_{\mathrm{flip}}.
\end{equation}
Here, $\mathcal{L}_{\mathrm{cls}}$ is a label-smoothed classification loss~\cite{label_smoothing},
$\mathcal{L}_{\mathrm{mim}}$ performs masked latent reconstruction following masked autoencoding~\cite{mae},
and $\mathcal{L}_{\mathrm{flip}}$ encourages spatial consistency under horizontal flipping.
These objectives respectively promote class-level semantics, local latent structure, and spatial equivariance in the dense representation.
Detailed formulations and implementation settings are provided in the appendix.

We use the composed dense features to supervise an intermediate SiT layer following the representation-alignment formulation of REPA~\cite{repa}.
Let
$\boldsymbol{G}^{(\ell)}
=
[\boldsymbol{g}^{(\ell)}_1,\ldots,\boldsymbol{g}^{(\ell)}_N]$
denote the $N$ patch-token features produced by the $\ell$-th SiT block.
Following REPA, a lightweight MLP projector $\Pi$ maps the SiT features into the composer feature space, and spatially corresponding tokens are aligned using cosine distance:
\begin{equation}
\mathcal{L}_{\mathrm{align}}
=
\frac{1}{N}
\sum_{i=1}^{N}
d_{\cos}
\left(
\Pi(\boldsymbol{g}^{(\ell)}_i),
\operatorname{sg}(\boldsymbol{h}_i)
\right),
\end{equation}
where $d_{\cos}(\cdot,\cdot)$ denotes cosine distance and
$\operatorname{sg}(\cdot)$ is the stop-gradient operator.
Thus, $\mathcal{L}_{\mathrm{align}}$ updates SiT and the projector, but does not back-propagate into the composer; the composer itself is optimized only by $\mathcal{L}_{\mathrm{comp}}$.

\subsection{Stage-Adaptive Supervision and Soft Label Conditioning}
\label{sec:stage}

Representation alignment is effective early in training but becomes detrimental when maintained for too long. CoVA therefore disables alignment after a transition step $s_{\tau}$ and reuses the composer's semantic prediction for sample-dependent class conditioning.

For a training sample $j$, the composer produces the detached posterior
\begin{equation}
\boldsymbol{q}_j
=
\operatorname{sg}\left(\operatorname{softmax}(\boldsymbol{o}_j)\right),
\qquad
\boldsymbol{q}_j\in\mathbb{R}^{K}.
\end{equation}
Unlike fixed label smoothing, $\boldsymbol{q}_j$ is sample-dependent and captures the composer's predicted relations among classes.

Let $\boldsymbol{E}\in\mathbb{R}^{K\times d_c}$ denote the SiT class-embedding table. With probability $p_s$, the hard-label condition is replaced by the naive soft condition
\begin{equation}
\bar{\boldsymbol{c}}_j
=
\boldsymbol{q}_j^{\top}\boldsymbol{E}.
\end{equation}
Since this interpolation changes both condition direction and magnitude, we restore its magnitude to that of the corresponding hard-label embedding:
\begin{equation}
\boldsymbol{c}^{\mathrm{soft}}_j
=
\frac{\|\boldsymbol{E}_{y_j}\|_2}{\|\bar{\boldsymbol{c}}_j\|_2+\epsilon}
\bar{\boldsymbol{c}}_j,
\label{eq:soft_label_condition}
\end{equation}
where $y_j$ is the ground-truth class and $\epsilon$ ensures numerical stability. This preserves the posterior-dependent direction while removing the unintended change in conditioning strength.

During training, each sample uses either $\boldsymbol{E}_{y_j}$ or
$\boldsymbol{c}^{\mathrm{soft}}_j$ according to $p_s$, followed by standard
classifier-free label dropout. Let $s_\tau$ and $s_\tau'$ denote the
alignment stopping and SLC activation steps, respectively; alignment is used
for $s < s_\tau$, while sample-dependent soft conditioning is activated for
$s \ge s_\tau'$.

The overall training objective is
\begin{equation}
\mathcal{L}
=
\mathcal{L}_{\mathrm{FM}}(\boldsymbol{c})
+
\lambda_{\mathrm{align}}(s)\mathcal{L}_{\mathrm{align}}
+
\lambda_{\mathrm{comp}}\mathcal{L}_{\mathrm{comp}}.
\end{equation}
Implementation details of the conditioning schedule are deferred to the appendix. At inference, the composer and alignment projector are discarded, and the model uses the standard SiT pipeline with hard-label conditioning.

\section{Experiments}

\subsection{Setup}
\label{sec:setup}
\noindent\textbf{Experimental setup.}
We conduct all experiments on ImageNet~\cite{imagenet} at $256\times256$ resolution, following ADM preprocessing~\cite{adm}. Images are encoded with the pretrained Stable Diffusion VAE~\cite{ldm}, yielding pre-extracted $4\times32\times32$ latent codes. Unless otherwise specified, we use SiT-L/2~\cite{sit} with $2\times2$ latent patches and train for $400$K iterations using SiT's linear flow-matching path and velocity parameterization. SiT and the composer are optimized with AdamW~\cite{adam2} at a constant learning rate of $10^{-4}$, zero weight decay, and global batch size $256$. We use EMA with decay $0.9999$ and mixed-precision training.

The CoVA composer is a tiny ViT~\cite{vit} block with width $192$, $12$ Transformer blocks, $3$ attention heads, and MLP ratio $4.0$, trained throughout the full $400$K iterations. We use label smoothing $0.1$, masking ratio $0.7$, MSE for masked latent reconstruction, cosine distance for flip consistency, and set all auxiliary loss weights to $1.0$. Following REPA~\cite{repa}, a token-wise MLP projector maps SiT features to the $192$-dimensional composer space. Features are extracted from blocks $4$, $8$, and $8$ for SiT-B/2, SiT-L/2, and SiT-XL/2, respectively. The projector is a three-layer MLP with dimensions \(D\!\rightarrow\!1152\!\rightarrow\!1152\!\rightarrow\!192\), with SiLU activations after the first two layers, where \(D\) is the hidden dimension of the corresponding SiT backbone. Alignment is used for the first $200$K iterations, after which it is disabled and the detached composer posterior is used for SLC during the last 50K steps with replacement probability $0.9$ and the magnitude correction from Sec.~\ref{sec:stage}. All methods share the same backbone, optimization schedule, training budget, and preprocessing unless stated otherwise.

\noindent\textbf{Generation quality assessment.}
We report FID~\cite{fid}, sFID~\cite{sfid}, IS~\cite{is}, Precision, and Recall~\cite{rec}. We generate $50$K samples from the EMA model using the SDE Euler--Maruyama solver with $250$ integration steps, and evaluate against ImageNet reference statistics. Classifier-free guidance~\cite{cfg} is disabled unless otherwise specified, and the same protocol is used for all methods.

\begin{figure*}[t]
\vspace{0.05in}

\begin{minipage}[t]{0.418\textwidth}
\centering\small
\captionof{table}{\textbf{FID comparison across training iterations for accelerated alignment methods.} All experiments are conducted on ImageNet (256$\times$256) with a batch size of 256 and without CFG.}
\resizebox{0.95\textwidth}{!}{%
\begin{tabular}{lccc}
\arrayrulecolor{black}\toprule
     Method  & \#Params & Iter. & FID$\downarrow$  \\
     \midrule
     SiT-B/2  &130M & 400K & 33.0 \\
     +REPA  &130M & 400K &  24.4 \\
     +SRA2  &130M & 400K & 28.9 \\
     {\textbf{+ \name (ours)}} &130M & \textbf{400K}  & \textbf{21.4}\\
     \arrayrulecolor{black!40}\midrule
     SiT-L/2 &458M & 400K & 18.8 \\
     +REPA &458M & 400K & 9.7 \\
     +SRA2 &458M & 200K & 22.5 \\
     +SRA2 &458M & 400K & 14.3 \\
     
     {\textbf{+ \name (ours)}} &458M & \textbf{200K}  & \textbf{20.4}\\
     {\textbf{+ \name (ours)}} &458M & \textbf{400K} & \textbf{9.3} \\
     \arrayrulecolor{black!40}\midrule
     SiT-XL/2  &675M & 400K   & 17.2\\
     +REPA  &675M & 400K   & 7.9\\
     +REPA  &675M & 1M   & 6.4\\
     +REPA  &675M & 4M   & 5.9\\
     +SRA2  &675M & 400K   & 11.7\\
     +SRA2  &675M & 1M   & 8.2\\
     +SRA2  &675M & 2M   & 7.0\\
     
     {\textbf{+ \name (ours)}} &675M & \textbf{400K}  & \textbf{7.4}\\
     {\textbf{+ \name (ours)}} &675M & \textbf{1M}  & \textbf{4.4}\\
     {\textbf{+ \name (ours)}} &675M & \textbf{2M}  & \textbf{3.6}\\
     \arrayrulecolor{black}\bottomrule
\end{tabular}
}
\label{tab:wo_cfg}
\end{minipage}
~~
\begin{minipage}[t]{0.53\textwidth}
\centering
\large
\captionof{table}{
    \textbf{Comparison with state-of-the-art methods on
    ImageNet 256$\times$256 with CFG.}
    ``External'' indicates reliance on additional pretrained
    models beyond the standard generative pipeline.
}
\vspace{-0.05in}

\resizebox{\textwidth}{!}{%
\begin{tabular}{l c c c c c c c}
\toprule
{\pz\pz Method}
& Epochs
& {\pz FID$\downarrow$}
& sFID$\downarrow$
& IS$\uparrow$
& Pre.$\uparrow$
& Rec.$\uparrow$
& External \\
\arrayrulecolor{black}\midrule

\multicolumn{8}{l}{
    \emph{Pixel diffusion}\vspace{0.02in}
} \\

\pz\pz ADM-U
& \pz400 & \pz3.94 & 6.14 & 186.7 & 0.82 & 0.52
& \extno \\

\pz\pz VDM$++$
& \pz560 & \pz2.40 & -- & 225.3 & -- & --
& \extno \\

\pz\pz Simple diffusion
& \pz800 & \pz2.77 & -- & 211.8 & -- & --
& \extno \\

\pz\pz CDM
& 2160 & \pz4.88 & -- & 158.7 & -- & --
& \extno \\

PixelFlow-XL/4
& 320 & 1.98 & -- & 282.1 & 0.81 & 0.60
& \extno \\

DiP-XL/16
& 320 & 1.98 & -- & 282.9 & 0.80 & 0.62
& \extno \\

PixNerd-XL/16
& 320 & 1.95 & -- & 298.0 & 0.80 & 0.60
& \extyes \\

PixelGen-XL/16
& 160 & 1.83 & -- & 293.6 & 0.79 & 0.63
& \extyes \\

JiT-G/16
& 600 & 1.82 & -- & 292.6 & -- & --
& \extno \\

DeCo-XL/16
& 800 & 1.62 & 4.41 & 301.0 & 0.80 & 0.62
& \extyes \\

\arrayrulecolor{black!40}\midrule

\multicolumn{8}{l}{
    \emph{Latent diffusion, Transformer + U-Net hybrid}
    \vspace{0.02in}
} \\

\pz\pz U-ViT-H/2
& \pz240 & \pz2.29 & 5.68 & 263.9 & 0.82 & 0.57
& \extno \\

\pz\pz DiffiT
& -- & \pz1.73 & -- & 276.5 & 0.80 & 0.62
& \extno \\

\pz\pz MDTv2-XL/2
& 1080 & \pz1.58 & 4.52 & 314.7 & 0.79 & 0.65
& \extno \\

\arrayrulecolor{black}\midrule

\multicolumn{8}{l}{
    \emph{Latent diffusion, Transformer}\vspace{0.02in}
} \\

\pz\pz MaskDiT
& 1600 & \pz2.28 & 5.67 & 276.6 & 0.80 & 0.61
& \extno \\

\pz\pz SD-DiT
& \pz480 & \pz3.23 & -- & -- & -- & --
& \extno \\

\arrayrulecolor{black!30}\cmidrule(lr){1-8}

\pz\pz DiT-XL/2
& 1400 & \pz2.27 & 4.60 & 278.2
& 0.83 & 0.57
& \extno \\

\arrayrulecolor{black!30}\cmidrule(lr){1-8}

\pz\pz SiT-XL/2
& 1400 & \pz2.06 & 4.50 & 270.3 & 0.82 & 0.59
& \extno \\

\pz\pz + REPA
& \pz800 & \pz1.42 & 4.70 & 305.7 & 0.80 & 0.65
& \extyes \\

\pz\pz + REG
& \pz800 & 1.36 & 4.25 & 299.4 & 0.77 & 0.66
& \extyes \\

\pz\pz + SRA2
& \pz200 & 1.98 & 4.53 & 284.5 & 0.83 & 0.59
& \extno \\

\pz\pz + SRA2
& \pz400 & 1.84 & 4.50 & 306.5 & 0.82 & 0.61
& \extno \\

\pz\pz \textbf{+ CoVA (ours)}
& 200 & 2.19 & 4.49 & 287.9 & 0.80 & 0.60
& \extno \\

\pz\pz \textbf{+ CoVA (ours)}
& 400 & 2.03 & 4.52 & 316.4& 0.80 & 0.62
& \extno \\

\arrayrulecolor{black}\bottomrule
\end{tabular}%
}

\label{tab:main}
\end{minipage}
\vspace{-10pt}
\end{figure*}

\subsection{System-Level Comparisons}
\label{sec:system_comparisons}
\noindent\textbf{Comparison with SOTA methods.}
Tables~\ref{tab:wo_cfg} and~\ref{tab:main} compare CoVA with representative generative models~\cite{pixelflow,chen2025dip,pixnerd,pixelgen,jit,ma2025deco} and representation-alignment methods on ImageNet $256\times256$, without and with classifier-free guidance, respectively. Under the same 400K-step budget, CoVA consistently improves FID across SiT-B/2, SiT-L/2, and SiT-XL/2, from 33.0 to 21.4, 18.8 to 9.3, and 17.2 to 7.4, respectively. With CFG, CoVA achieves an FID of 2.03 and an IS of 316.4 in only 400 epochs, compared with 2.06 and 270.3 for vanilla SiT-XL/2 after 1,400 epochs. These results suggest that CoVA's largest advantage
appears in improving the underlying unguided generative model, while strong
classifier-free guidance partially narrows the differences among training
methods.

\begin{wraptable}{r}{0.48\linewidth}
  \vspace{-1.2em}
  \centering
  \captionsetup{font={footnotesize}}
  \setlength{\tabcolsep}{3pt}
  \caption{
    \textbf{Training-time computational overhead on SiT-XL/2.}
    \ EFP denotes additional feature-processing parameters, reported as
    \emph{\textcolor{Rhodamine}{external model parameters} + \textcolor{cyan}{trainable auxiliary parameters}}.
}

  \vspace{-3pt}
  \scriptsize
  \resizebox{1.0\linewidth}{!}{%
  \begin{tabular}{l|cccc}
      \hline
      Method
      & \#EFP (M)
      & TS (Iter/s)$\uparrow$
      & GFLOPs$\downarrow$
      & Latency (ms)$\downarrow$ \\
      \hline
      SiT-XL/2
      & \textcolor{Rhodamine}{0}+\textcolor{cyan}{0}
      & 8.1
      & 114.46 
      & 18.05 \\

      +REPA
      & \textcolor{Rhodamine}{86}+\textcolor{cyan}{8}
      & 6.3~(\textcolor{red}{-22\%})
      & 138.50~(\textcolor{red}{+21\%})
      & 22.81~(\textcolor{red}{+26\%}) \\

      +SRA2
      & \textcolor{Rhodamine}{0}+\textcolor{cyan}{18}
      & 7.2~(\textcolor{red}{\textbf{-11\%}})
      & 118.55~(\textcolor{red}{\textbf{+4\%}})
      & 19.20~(\textcolor{red}{\textbf{+6\%}}) \\

      \hline
      +\name
      & \textcolor{Rhodamine}{0}+\textcolor{cyan}{8.5}
      & 7.0~(\textcolor{red}{\textbf{-14\%}})
      & 122.69~(\textcolor{red}{\textbf{+7.2\%}})
      & 21.17~(\textcolor{red}{\textbf{+17\%}}) \\
      \hline
  \end{tabular}
  }
  \vspace{-1em}
  \label{tab:flops}
\end{wraptable}

\noindent\textbf{Training computational cost.}
\noindent\textbf{Training-time computational overhead.}
CoVA avoids the external pretrained representation encoder used by methods such as REPA and instead introduces a lightweight online latent composer. As shown in Table~\ref{tab:flops}, CoVA requires no external model and introduces only $8.5$M trainable auxiliary parameters, substantially fewer than the $94$M additional parameters required by REPA and the $18$M auxiliary parameters used by SRA2. On SiT-XL/2, this increases the training computation from $114.46$ to $122.69$ GFLOPs ($+7.2\%$), reduces training throughput from $8.1$ to $7.0$ iter/s ($-14\%$), and increases latency from $18.05$ to $21.17$ ms ($+17\%$). This overhead remains substantially lower than that of REPA, which relies on a large external encoder. Importantly, all CoVA-specific modules are used only during training and are discarded at inference, leaving the original SiT sampling pipeline unchanged.

\begin{figure*}[t]
    \centering
    \includegraphics[width=1.0\linewidth]{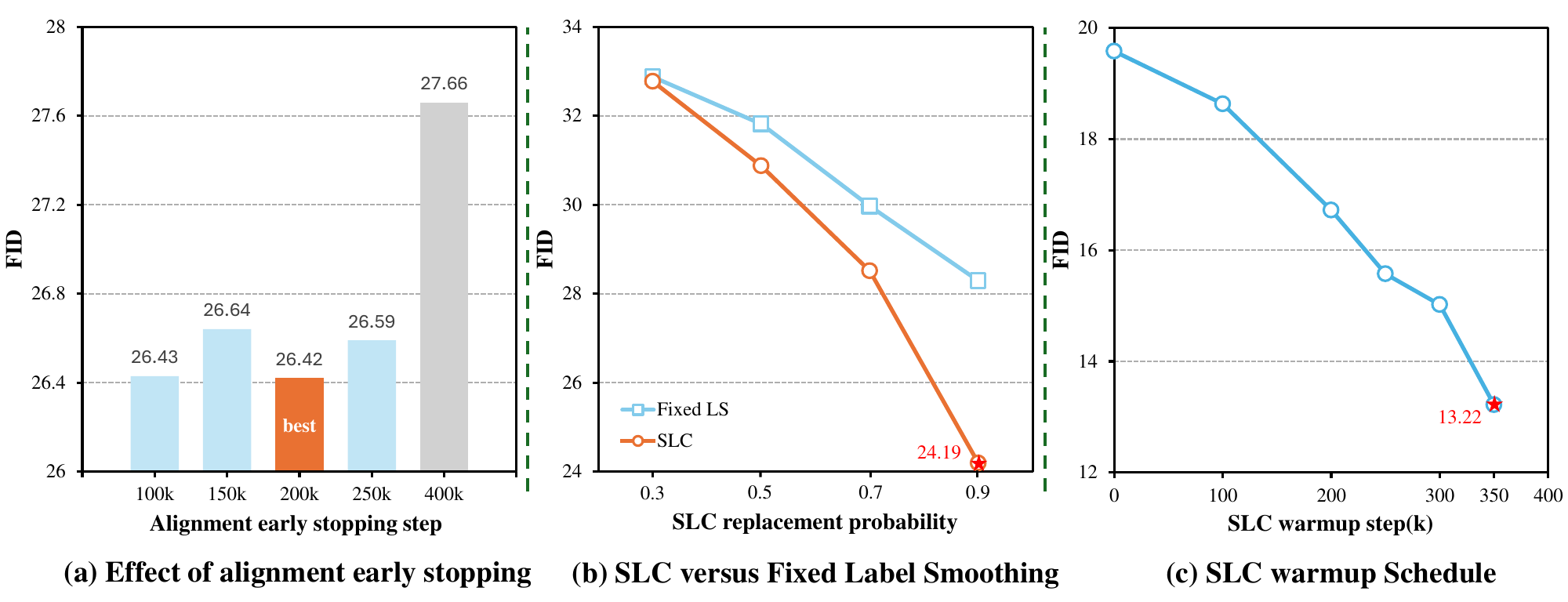}
    \caption{\textbf{Stage-wise scheduling of representation alignment and conditioning.} (a) Early stopping consistently outperforms full-training alignment, with stable performance across a broad stopping range. (b) Sample-adaptive SLC consistently outperforms fixed label smoothing across all replacement probabilities, with the advantage becoming more pronounced as the replacement probability increases. (c) Longer naive SLC warmup consistently improves performance across the tested schedules. For both (a) and (b), the MIM mask ratio is fixed to $0.5$.}  
    \label{fig:ablation}
    \vspace{-10pt}
\end{figure*}


\subsection{Ablation Studies}
\label{sec:ablation}

All ablations use SiT-B/2 trained for 400k steps with 250 SDE function evaluations and without CFG.

\begin{wraptable}[12]{r}{0.50\linewidth}
\vspace{-12pt}
\centering
\captionsetup{font={footnotesize}}
\setlength{\tabcolsep}{3pt}
\caption{
\textbf{Ablation on composer training objectives.}
All variants use the same feature-alignment configuration without SLC or
early stopping. When enabled, classification uses label smoothing
$\epsilon=0.1$, and MIM uses a masking ratio of $0.7$.
Prec.\ and Rec.\ denote precision and recall.
}
\vspace{-3pt}
\scriptsize
\resizebox{0.95\linewidth}{!}{
\begin{tabular}{ccc|ccccc}
\hline
$\mathcal{L}_{\mathrm{cls}}$ &
$\mathcal{L}_{\mathrm{mim}}$ &
$\mathcal{L}_{\mathrm{flip}}$
& FID$\downarrow$ & sFID$\downarrow$
& IS$\uparrow$ & Prec.$\uparrow$ & Rec.$\uparrow$ \\
\hline
$\times$ & $\times$ & $\times$
& 33.02 & 6.46 & 43.71 & 0.53 & 0.63 \\
$\times$ & $\checkmark$ & $\times$
& 31.92 & 6.26 & 48.14 & 0.54 & \textbf{0.64} \\
$\checkmark$ & $\times$ & $\times$
& 30.85 & 6.83 & 50.94 & 0.54 & 0.63 \\
$\checkmark$ & $\checkmark$ & $\times$
& 28.11 & 6.41 & 54.91 & 0.56 & \textbf{0.64} \\
$\checkmark$ & $\times$ & $\checkmark$
& 28.06 & \textbf{6.23} & 54.77 & 0.56 & 0.62 \\
$\checkmark$ & $\checkmark$ & $\checkmark$
& \textbf{26.83} & 6.32 & \textbf{57.69}
& \textbf{0.57} & \textbf{0.64} \\
\hline
\end{tabular}
}
\label{tab:ablation_Composer_objectives}
\vspace{-5em}
\end{wraptable}

\noindent
{\setlength{\fboxsep}{1pt}%
\colorbox{green!20}{\textbf{Composer Objectives.}}}%
\hspace{0.2em}%
We first isolate the contribution of each objective used to train the latent composer. As shown in Table~\ref{tab:ablation_Composer_objectives}, masked latent reconstruction alone improves FID from $33.02$ to $31.92$, while classification supervision alone further reduces it to $30.85$. Starting from classification supervision, adding masked reconstruction or horizontal-flip consistency yields comparable improvements, and combining all three objectives achieves the best FID of $26.83$, together with the highest IS and Precision. These results suggest that the objectives provide complementary supervision: classification organizes semantic information, while masked reconstruction and flip consistency preserve instance-level and spatial structure in the dense representations.

\vspace{0.4em}

\noindent
{\setlength{\fboxsep}{1pt}%
\colorbox{yellow!20}{\textbf{Alignment Early Stopping.}}}%
\hspace{0.2em}%
We next investigate how long the composed representation should be used as an alignment target. As shown in Fig.~\ref{fig:ablation}(a), early stopping consistently outperforms full-training alignment, with stable performance across a broad stopping range: stopping alignment between $100$K and $250$K iterations yields FIDs between $26.42$ and $26.64$, compared with $27.66$ when alignment remains active throughout training. We therefore stop the representation-alignment objective at $200$K iterations in the subsequent experiments.

\vspace{0.4em}

\noindent
{\setlength{\fboxsep}{1pt}%
\colorbox{blue!15}{\textbf{SLC versus Fixed Label Smoothing.}}}%
\hspace{0.2em}%
We then ask whether the benefit of soft conditioning comes from generic label smoothing or from the sample-dependent posterior produced by the online classifier. Under matched replacement probabilities, sample-adaptive SLC consistently outperforms fixed label smoothing, and the gap becomes more pronounced as the replacement probability increases, reaching $24.19$ versus $28.29$ FID at $p=0.9$ in Fig.~\ref{fig:ablation}(b). This result shows that the gain is not explained by weakening the hard condition alone. The instance-specific posterior provides additional useful information for generation. Since sampling uses hard-label conditioning, we cap the SLC replacement probability at $p=0.9$ to retain sufficient hard-label exposure during training and limit the train--test mismatch.

\vspace{0.4em}

\noindent
{\setlength{\fboxsep}{1pt}%
\colorbox{orange!20}{\textbf{SLC Warmup Schedule.}}}%
\hspace{0.2em}%
Finally, we study when the generator should start receiving the online classifier's posterior. As shown in Fig.~\ref{fig:ablation}(c), delaying SLC consistently improves performance across the tested schedules. We attribute this trend to the stage-dependent reliability of the posterior: early in training, the composer is still evolving and its predictions may provide unstable semantic mixtures that interfere with the formation of a robust hard-label-conditioned generative trajectory. Once the composer has matured, its posterior becomes more suitable as a sample-adaptive refinement of the class condition. We therefore enable SLC only during the final $50$K iterations.

\subsection{Understanding Soft Label Conditioning:
Disentangling Direction and Magnitude}
\label{sec:slc_analysis}

\begin{table}[t]
\centering

\begin{minipage}[t]{0.54\textwidth}
\vspace{0pt}
\centering
\captionsetup{font={footnotesize}}
\setlength{\tabcolsep}{3pt}

\captionof{table}{
\textbf{Direction--magnitude decomposition of SLC.}
Direction and magnitude refer to the training-time class condition.
All variants use standard hard-label conditioning at inference.
}

\label{tab:slc_geometry}

\vspace{-3pt}
\scriptsize
\resizebox{\linewidth}{!}{%
\begin{tabular}{ll|ccccc}
\hline
Direction & Magnitude
& FID$\downarrow$ & sFID$\downarrow$
& IS$\uparrow$ & Prec.$\uparrow$ & Rec.$\uparrow$ \\
\hline
Hard & Hard
& 24.87 & 5.95 & 61.19 & 0.58 & 0.64 \\

Soft & Soft
& \textbf{12.71} & 5.58 & \textbf{118.34}
& 0.67 & 0.58 \\

Hard & Soft
& 12.78 & \textbf{5.55} & 109.71
& \textbf{0.68} & 0.57 \\

Soft & Hard
& 22.73 & 5.77 & 70.30
& 0.58 & 0.63 \\

Shuffled soft & Hard
& 34.13 & 6.47 & 42.35
& 0.52 & \textbf{0.64} \\
\hline
\end{tabular}%
}
\end{minipage}
\hfill
\begin{minipage}[t]{0.44\textwidth}
\vspace{0pt}
\centering
\captionsetup{font={footnotesize}}
\setlength{\tabcolsep}{3pt}

\captionof{table}{
\textbf{Sensitivity to inference-time condition magnitude.}
We scale hard-label embeddings by $\gamma$ at inference for a model trained without SLC, keeping all parameters fixed.
}
\label{tab:condition_scaling}

\vspace{-3pt}
\scriptsize
\resizebox{\linewidth}{!}{%
\begin{tabular}{c|ccccc}
\hline
$\gamma$
& FID$\downarrow$ & sFID$\downarrow$
& IS$\uparrow$ & Prec.$\uparrow$ & Rec.$\uparrow$ \\
\hline
1.00
& 24.87 & 5.95 & 61.19
& 0.58 & \textbf{0.64} \\

1.10
& 18.42 & 5.74 & 80.46
& 0.63 & 0.61 \\

1.20
& 13.80 & 5.62 & 102.67
& 0.67 & 0.58 \\

1.35
& \textbf{9.38} & \textbf{5.60}
& \textbf{137.38} & \textbf{0.71} & 0.55 \\
\hline
\end{tabular}%
}
\end{minipage}

\vspace{-0.8em}
\end{table}

Naive Soft Label Conditioning (SLC) averages class embeddings according to the
sample-dependent composer posterior, changing both the \emph{direction} and
\emph{magnitude} of the condition. For ground-truth class $y$, let
$\boldsymbol{e}_{h}=\boldsymbol{e}_{y}$ and
\begin{equation}
\boldsymbol{e}_{s}=\sum_k q_k \boldsymbol{e}_k
\end{equation}
denote the hard and naive soft conditions, respectively. We disentangle their
effects by constructing
\begin{equation}
\boldsymbol{e}_{\mathrm{hard\text{-}dir,soft\text{-}mag}}
=\|\boldsymbol{e}_{s}\|_2
\frac{\boldsymbol{e}_{h}}{\|\boldsymbol{e}_{h}\|_2},
\qquad
\boldsymbol{e}_{\mathrm{soft\text{-}dir,hard\text{-}mag}}
=\|\boldsymbol{e}_{h}\|_2
\frac{\boldsymbol{e}_{s}}{\|\boldsymbol{e}_{s}\|_2}.
\label{eq:slc_decomposition}
\end{equation}

\textbf{Direction and magnitude play different roles.}
As shown in Table~\ref{tab:slc_geometry}, using the soft magnitude alone reduces FID
from 24.87 to 12.78, but also lowers Recall from 0.64 to 0.57. By contrast, using the soft direction with the hard-label magnitude gives FID improvement to 22.73 while preserving
higher Recall. Importantly, shuffling soft directions across samples degrades
FID sharply to 34.13, showing that the directional component carries
sample-specific information rather than generic smoothing. Thus, the modest
standalone FID gain of direction should not be interpreted as semantic
irrelevance, preserving the correspondence between each sample and its
posterior direction is critical. This suggests that the two components encode
qualitatively different effects: magnitude mainly changes conditioning
strength, whereas direction reflects posterior-dependent class structure.

\textbf{Magnitude creates a conditioning-strength shortcut.}
Using the attenuated soft magnitude during training while reverting to the
standard hard-label condition at inference introduces an implicit
conditioning-strength mismatch. This alone produces a large change in the
fidelity--coverage balance, even without introducing sample-dependent
semantic direction. The inference-time scaling experiment in Table~\ref{tab:condition_scaling} independently confirms
that the generator is highly sensitive to this coordinate: modifying the
hard-label condition magnitude substantially changes FID, Precision, and
Recall while keeping all model parameters fixed. Therefore, the large gain
of naive SLC cannot be attributed solely to semantic interpolation.

\section{Conclusion}

We presented \textbf{CoVA}, a lightweight framework that reorganizes VAE representations and reuses them adaptively throughout flow-matching training. A small online composer first transforms reconstruction-oriented VAE features into more effective targets for intermediate representation alignment. As training progresses and explicit alignment becomes less beneficial, CoVA instead reuses the composer's class posterior for sample-dependent soft conditioning. We further show that naive soft label conditioning entangles semantic interpolation with changes in embedding magnitude, and introduce a simple magnitude correction to remove this confounding effect. CoVA requires no additional pretrained visual encoder and introduces no inference-time component.

\subsection*{AI use statement}
Generative AI tools were used for manuscript polishing, literature retrieval and discovery, and code review. All experimental implementations, results, analyses, and scientific claims were independently verified by the authors.

\subsection*{Reproducibility statement}
We provide detailed descriptions of the CoVA architecture, training
objectives, stage-adaptive training schedule, and soft label conditioning
strategy. Experimental settings, including model
architectures, optimization hyperparameters, alignment layers, auxiliary
loss configurations, and sampling protocols, are reported in
Sec.~\ref{sec:setup} and the Appendix.

\bibliography{iclr2027_conference}

\begin{thebibliography}{51}
\providecommand{\natexlab}[1]{#1}
\providecommand{\url}[1]{\texttt{#1}}
\expandafter\ifx\csname urlstyle\endcsname\relax
  \providecommand{\doi}[1]{doi: #1}\else
  \providecommand{\doi}{doi: \begingroup \urlstyle{rm}\Url}\fi

\bibitem[Bao et~al.()Bao, Nie, Xue, Cao, Li, Su, and Zhu]{u-vit}
Fan Bao, Shen Nie, Kaiwen Xue, Yue Cao, Chongxuan Li, Hang Su, and Jun Zhu.
\newblock All are worth words: A vit backbone for diffusion models.
\newblock In \emph{CVPR}.

\bibitem[Caron et~al.(2021)Caron, Touvron, Misra, J{\'e}gou, Mairal, Bojanowski, and Joulin]{dino}
Mathilde Caron, Hugo Touvron, Ishan Misra, Herv{\'e} J{\'e}gou, Julien Mairal, Piotr Bojanowski, and Armand Joulin.
\newblock Emerging properties in self-supervised vision transformers.
\newblock In \emph{2021 IEEE/CVF international conference on computer vision (ICCV)}, pp.\  9630--9640. IEEE, 2021.

\bibitem[Chen et~al.(2024)Chen, Yu, Ge, Yao, Xie, Wang, Kwok, Luo, Lu, and Li]{pixartalpha}
Junsong Chen, Jincheng Yu, Chongjian Ge, Lewei Yao, Enze Xie, Zhongdao Wang, James Kwok, Ping Luo, Huchuan Lu, and Zhenguo Li.
\newblock Pixart-$\alpha$: Fast training of diffusion transformer for photorealistic text-to-image synthesis.
\newblock In \emph{International conference on learning representations}, volume 2024, pp.\  57611--57640, 2024.

\bibitem[Chen et~al.(2025{\natexlab{a}})Chen, Cai, Chen, Xie, Yang, Tang, Li, and Han]{dc-ae}
Junyu Chen, Han Cai, Junsong Chen, Enze Xie, Shang Yang, Haotian Tang, Muyang Li, and Song Han.
\newblock Deep compression autoencoder for efficient high-resolution diffusion models.
\newblock In \emph{International Conference on Learning Representations}, volume 2025, pp.\  96539--96560, 2025{\natexlab{a}}.

\bibitem[Chen et~al.(2025{\natexlab{b}})Chen, Zou, He, Chen, Xie, Han, and Cai]{dc-ae2}
Junyu Chen, Dongyun Zou, Wenkun He, Junsong Chen, Enze Xie, Song Han, and Han Cai.
\newblock Dc-ae 1.5: Accelerating diffusion model convergence with structured latent space.
\newblock In \emph{2025 IEEE/CVF International Conference on Computer Vision (ICCV)}, pp.\  1--10. IEEE, 2025{\natexlab{b}}.

\bibitem[Chen et~al.(2025{\natexlab{c}})Chen, Ge, Zhang, Sun, and Luo]{pixelflow}
Shoufa Chen, Chongjian Ge, Shilong Zhang, Peize Sun, and Ping Luo.
\newblock Pixelflow: Pixel-space generative models with flow.
\newblock \emph{arXiv preprint arXiv:2504.07963}, 2025{\natexlab{c}}.

\bibitem[Chen et~al.(2025{\natexlab{d}})Chen, Zhu, Chen, Zhang, Hu, Zhao, Wang, Yang, and Tai]{chen2025dip}
Zhennan Chen, Junwei Zhu, Xu~Chen, Jiangning Zhang, Xiaobin Hu, Hanzhen Zhao, Chengjie Wang, Jian Yang, and Ying Tai.
\newblock Dip: Taming diffusion models in pixel space.
\newblock \emph{arXiv preprint arXiv:2511.18822}, 2025{\natexlab{d}}.

\bibitem[Deng et~al.(2009)Deng, Dong, Socher, et~al.]{imagenet}
Jia Deng, Wei Dong, Richard Socher, et~al.
\newblock Imagenet: A large-scale hierarchical image database.
\newblock In \emph{CVPR}, 2009.

\bibitem[Dhariwal \& Nichol(2021)Dhariwal and Nichol]{adm}
Prafulla Dhariwal and Alexander Nichol.
\newblock Diffusion models beat gans on image synthesis.
\newblock In \emph{NeurIPS}, 2021.

\bibitem[Dosovitskiy et~al.(2020)Dosovitskiy, Beyer, Kolesnikov, Weissenborn, Zhai, Unterthiner, Dehghani, Minderer, Heigold, Gelly, et~al.]{vit}
Alexey Dosovitskiy, Lucas Beyer, Alexander Kolesnikov, Dirk Weissenborn, Xiaohua Zhai, Thomas Unterthiner, Mostafa Dehghani, Matthias Minderer, Georg Heigold, Sylvain Gelly, et~al.
\newblock An image is worth 16x16 words: Transformers for image recognition at scale.
\newblock \emph{arXiv preprint arXiv:2010.11929}, 2020.

\bibitem[Esser et~al.(2024)Esser, Kulal, Blattmann, Entezari, M{\"u}ller, Saini, Levi, Lorenz, Sauer, Boesel, et~al.]{sd3}
Patrick Esser, Sumith Kulal, Andreas Blattmann, Rahim Entezari, Jonas M{\"u}ller, Harry Saini, Yam Levi, Dominik Lorenz, Axel Sauer, Frederic Boesel, et~al.
\newblock Scaling rectified flow transformers for high-resolution image synthesis.
\newblock In \emph{Forty-first international conference on machine learning}, 2024.

\bibitem[Gao et~al.(2023)Gao, Zhou, Cheng, et~al.]{mdtv2}
Shanghua Gao, Pan Zhou, Ming-Ming Cheng, et~al.
\newblock Mdtv2: Masked diffusion transformer is a strong image synthesizer.
\newblock In \emph{arXiv preprint arXiv:2303.14389}, 2023.

\bibitem[Hatamizadeh et~al.()Hatamizadeh, Song, Liu, Kautz, and Vahdat]{diffit}
Ali Hatamizadeh, Jiaming Song, Guilin Liu, Jan Kautz, and Arash Vahdat.
\newblock Diffit: Diffusion vision transformers for image generation.
\newblock In \emph{ECCV}.

\bibitem[He et~al.(2022)He, Chen, Xie, Li, Doll{\'a}r, and Girshick]{mae}
Kaiming He, Xinlei Chen, Saining Xie, Yanghao Li, Piotr Doll{\'a}r, and Ross Girshick.
\newblock Masked autoencoders are scalable vision learners.
\newblock In \emph{Proceedings of the IEEE/CVF Conference on Computer Vision and Pattern Recognition}, pp.\  16000--16009, 2022.

\bibitem[Heusel et~al.(2017)Heusel, Ramsauer, Unterthiner, et~al.]{fid}
Martin Heusel, Hubert Ramsauer, Thomas Unterthiner, et~al.
\newblock Gans trained by a two time-scale update rule converge to a local nash equilibrium.
\newblock In \emph{NeurIPS}, 2017.

\bibitem[Ho \& Salimans(2022)Ho and Salimans]{cfg}
Jonathan Ho and Tim Salimans.
\newblock Classifier-free diffusion guidance.
\newblock \emph{arXiv preprint arXiv:2207.12598}, 2022.

\bibitem[Ho et~al.(2020)Ho, Jain, and Abbeel]{ddpm}
Jonathan Ho, Ajay Jain, and Pieter Abbeel.
\newblock Denoising diffusion probabilistic models.
\newblock In \emph{NeurIPS}, 2020.

\bibitem[Hoogeboom et~al.()Hoogeboom, Heek, and Salimans]{simple}
Emiel Hoogeboom, Jonathan Heek, and Tim Salimans.
\newblock Simple diffusion: End-to-end diffusion for high resolution images.
\newblock In \emph{ICML}.

\bibitem[Jiang et~al.(2025)Jiang, Wang, Li, Zhang, Wang, Wei, Dai, Zhang, and Wang]{sra}
Dengyang Jiang, Mengmeng Wang, Liuzhuozheng Li, Lei Zhang, Haoyu Wang, Wei Wei, Guang Dai, Yanning Zhang, and Jingdong Wang.
\newblock No other representation component is needed: Diffusion transformers can provide representation guidance by themselves.
\newblock \emph{arXiv preprint arXiv:2505.02831}, 2025.

\bibitem[Kingma \& Gao()Kingma and Gao]{vdm}
Diederik Kingma and Ruiqi Gao.
\newblock Understanding diffusion objectives as the elbo with simple data augmentation.
\newblock In \emph{NeurIPS}.

\bibitem[Kingma \& Welling(2013)Kingma and Welling]{vae}
Diederik~P Kingma and Max Welling.
\newblock Auto-encoding variational bayes.
\newblock \emph{arXiv preprint arXiv:1312.6114}, 2013.

\bibitem[Kynk{\"a}{\"a}nniemi et~al.(2019)Kynk{\"a}{\"a}nniemi, Karras, Laine, et~al.]{rec}
Tuomas Kynk{\"a}{\"a}nniemi, Tero Karras, Samuli Laine, et~al.
\newblock Improved precision and recall metric for assessing generative models.
\newblock In \emph{NeurIPS}, 2019.

\bibitem[Leng et~al.(2025)Leng, Singh, Hou, et~al.]{repa-e}
Xingjian Leng, Jaskirat Singh, Yunzhong Hou, et~al.
\newblock Repa-e: Unlocking vae for end-to-end tuning with latent diffusion transformers.
\newblock In \emph{arXiv preprint arXiv:2504.10483}, 2025.

\bibitem[Li \& He(2025)Li and He]{jit}
Tianhong Li and Kaiming He.
\newblock Back to basics: Let denoising generative models denoise.
\newblock \emph{arXiv preprint arXiv:2511.13720}, 2025.

\bibitem[Lipman et~al.(2022)Lipman, Chen, Ben-Hamu, Nickel, and Le]{fm}
Yaron Lipman, Ricky~TQ Chen, Heli Ben-Hamu, Maximilian Nickel, and Matt Le.
\newblock Flow matching for generative modeling.
\newblock \emph{arXiv preprint arXiv:2210.02747}, 2022.

\bibitem[Loshchilov \& Hutter(2017)Loshchilov and Hutter]{adam2}
Ilya Loshchilov and Frank Hutter.
\newblock Decoupled weight decay regularization.
\newblock \emph{arXiv preprint arXiv:1711.05101}, 2017.

\bibitem[Ma et~al.(2024)Ma, Goldstein, Albergo, et~al.]{sit}
Nanye Ma, Mark Goldstein, Michael~S Albergo, et~al.
\newblock Sit: Exploring flow and diffusion-based generative models with scalable interpolant transformers.
\newblock In \emph{ECCV}, 2024.

\bibitem[Ma et~al.(2025{\natexlab{a}})Ma, Yu, Liu, Fang, and Wang]{ssd}
Xinyin Ma, Runpeng Yu, Songhua Liu, Gongfan Fang, and Xinchao Wang.
\newblock Diffusion model is effectively its own teacher.
\newblock In \emph{2025 IEEE/CVF Conference on Computer Vision and Pattern Recognition (CVPR)}, pp.\  12901--12911. IEEE, 2025{\natexlab{a}}.

\bibitem[Ma et~al.(2025{\natexlab{b}})Ma, Wei, Wang, Zhang, and Tian]{ma2025deco}
Zehong Ma, Longhui Wei, Shuai Wang, Shiliang Zhang, and Qi~Tian.
\newblock Deco: Frequency-decoupled pixel diffusion for end-to-end image generation.
\newblock \emph{arXiv preprint arXiv:2511.19365}, 2025{\natexlab{b}}.

\bibitem[Ma et~al.(2026)Ma, Xu, and Zhang]{pixelgen}
Zehong Ma, Ruihan Xu, and Shiliang Zhang.
\newblock Pixelgen: Pixel diffusion beats latent diffusion with perceptual loss.
\newblock \emph{arXiv preprint arXiv:2602.02493}, 2026.

\bibitem[Nash et~al.(2021)Nash, Menick, Dieleman, et~al.]{sfid}
Charlie Nash, Jacob Menick, Sander Dieleman, et~al.
\newblock Generating images with sparse representations.
\newblock In \emph{arXiv preprint arXiv:2103.03841}, 2021.

\bibitem[Peebles \& Xie(2023)Peebles and Xie]{dit}
William Peebles and Saining Xie.
\newblock Scalable diffusion models with transformers.
\newblock In \emph{ICCV}, 2023.

\bibitem[Qin et~al.(2025)Qin, Zhuo, Xin, Du, Li, Fu, Lu, Li, Liu, Zhu, et~al.]{lumina2}
Qi~Qin, Le~Zhuo, Yi~Xin, Ruoyi Du, Zhen Li, Bin Fu, Yiting Lu, Xinyue Li, Dongyang Liu, Xiangyang Zhu, et~al.
\newblock Lumina-image 2.0: A unified and efficient image generative framework.
\newblock In \emph{2025 IEEE/CVF International Conference on Computer Vision (ICCV)}, pp.\  20031--20042. IEEE, 2025.

\bibitem[Radford et~al.(2021)Radford, Kim, Hallacy, Ramesh, Goh, Agarwal, Sastry, Askell, Mishkin, Clark, et~al.]{clip}
Alec Radford, Jong~Wook Kim, Chris Hallacy, Aditya Ramesh, Gabriel Goh, Sandhini Agarwal, Girish Sastry, Amanda Askell, Pamela Mishkin, Jack Clark, et~al.
\newblock Learning transferable visual models from natural language supervision.
\newblock In \emph{International conference on machine learning}, pp.\  8748--8763. PmLR, 2021.

\bibitem[Rombach et~al.(2022)Rombach, Blattmann, Lorenz, et~al.]{ldm}
Robin Rombach, Andreas Blattmann, Dominik Lorenz, et~al.
\newblock High-resolution image synthesis with latent diffusion models.
\newblock In \emph{CVPR}, 2022.

\bibitem[Salimans et~al.(2016)Salimans, Goodfellow, Zaremba, et~al.]{is}
Tim Salimans, Ian Goodfellow, Wojciech Zaremba, et~al.
\newblock Improved techniques for training gans.
\newblock In \emph{NeurIPS}, 2016.

\bibitem[Singh et~al.(2026)Singh, Leng, Wu, Zheng, Zhang, Shechtman, and Xie]{irepa}
Jaskirat Singh, Xingjian Leng, Zongze Wu, Liang Zheng, Richard Zhang, Eli Shechtman, and Saining Xie.
\newblock What matters for representation alignment: Global information or spatial structure?
\newblock In \emph{International Conference on Learning Representations}, 2026.

\bibitem[Song et~al.(2020)Song, Meng, and Ermon]{ddim}
Jiaming Song, Chenlin Meng, and Stefano Ermon.
\newblock Denoising diffusion implicit models.
\newblock In \emph{arXiv preprint arXiv:2010.02502}, 2020.

\bibitem[Szegedy et~al.(2016)Szegedy, Vanhoucke, Ioffe, Shlens, and Wojna]{label_smoothing}
Christian Szegedy, Vincent Vanhoucke, Sergey Ioffe, Jon Shlens, and Zbigniew Wojna.
\newblock Rethinking the inception architecture for computer vision.
\newblock In \emph{Proceedings of the IEEE Conference on Computer Vision and Pattern Recognition}, pp.\  2818--2826, 2016.

\bibitem[Vaswani et~al.(2017)Vaswani, Shazeer, Parmar, Uszkoreit, Jones, Gomez, Kaiser, and Polosukhin]{attention}
Ashish Vaswani, Noam Shazeer, Niki Parmar, Jakob Uszkoreit, Llion Jones, Aidan~N Gomez, {\L}ukasz Kaiser, and Illia Polosukhin.
\newblock Attention is all you need.
\newblock \emph{Advances in neural information processing systems}, 30, 2017.

\bibitem[Wang et~al.(2026{\natexlab{a}})Wang, Jiang, Li, Lin, Shen, Kong, Liu, Dai, and Wang]{sra2}
Mengmeng Wang, Dengyang Jiang, Liuzhuozheng Li, Yucheng Lin, Guojiang Shen, Xiangjie Kong, Yong Liu, Guang Dai, and Jingdong Wang.
\newblock Sra 2: Variational autoencoder self-representation alignment for efficient diffusion training.
\newblock \emph{arXiv preprint arXiv:2601.17830}, 2026{\natexlab{a}}.

\bibitem[Wang et~al.(2025)Wang, Gao, Zhu, Huang, and Wang]{pixnerd}
Shuai Wang, Ziteng Gao, Chenhui Zhu, Weilin Huang, and Limin Wang.
\newblock Pixnerd: Pixel neural field diffusion.
\newblock \emph{arXiv preprint arXiv:2507.23268}, 2025.

\bibitem[Wang et~al.(2026{\natexlab{b}})Wang, Zhao, Zhou, Li, Liang, Shi, Zhao, Zhou, Zhang, Wang, et~al.]{haste}
Ziqiao Wang, Wangbo Zhao, Yuhao Zhou, Zekai Li, Zhiyuan Liang, Mingjia Shi, Xuanlei Zhao, Pengfei Zhou, Kaipeng Zhang, Zhangyang Wang, et~al.
\newblock Repa works until it doesn’t: Early-stopped, holistic alignment supercharges diffusion training.
\newblock \emph{Advances in Neural Information Processing Systems}, 38:\penalty0 136854--136887, 2026{\natexlab{b}}.

\bibitem[Wu et~al.(2025)Wu, Zhang, Shi, Gao, Chen, Wang, Chen, Gao, Tang, Yang, et~al.]{reg}
Ge~Wu, Shen Zhang, Ruijing Shi, Shanghua Gao, Zhenyuan Chen, Lei Wang, Zhaowei Chen, Hongcheng Gao, Yao Tang, Jian Yang, et~al.
\newblock Representation entanglement for generation: Training diffusion transformers is much easier than you think.
\newblock \emph{arXiv preprint arXiv:2507.01467}, 2025.

\bibitem[Xu et~al.(2025)Xu, Yue, Wang, Teng, Zhang, Liu, Zhou, Ouyang, and Bai]{xu2025exploring}
Wanghan Xu, Xiaoyu Yue, Zidong Wang, Yao Teng, Wenlong Zhang, Xihui Liu, Luping Zhou, Wanli Ouyang, and Lei Bai.
\newblock Exploring representation-aligned latent space for better generation.
\newblock \emph{arXiv preprint arXiv:2502.00359}, 2025.

\bibitem[Yao et~al.(2025)Yao, Yang, and Wang]{vavae}
Jingfeng Yao, Bin Yang, and Xinggang Wang.
\newblock Reconstruction vs. generation: Taming optimization dilemma in latent diffusion models.
\newblock In \emph{arXiv preprint arXiv:2501.01423}, 2025.

\bibitem[Yu et~al.(2024)Yu, Kwak, Jang, et~al.]{repa}
Sihyun Yu, Sangkyung Kwak, Huiwon Jang, et~al.
\newblock Representation alignment for generation: Training diffusion transformers is easier than you think.
\newblock In \emph{arXiv preprint arXiv:2410.06940}, 2024.

\bibitem[Yue et~al.(2024)Yue, Wang, Lu, Sun, Wei, Ouyang, Bai, and Zhou]{dod}
Xiaoyu Yue, Zidong Wang, Zeyu Lu, Shuyang Sun, Meng Wei, Wanli Ouyang, Lei Bai, and Luping Zhou.
\newblock Diffusion models need visual priors for image generation.
\newblock \emph{arXiv preprint arXiv:2410.08531}, 2024.

\bibitem[Zheng et~al.(2026)Zheng, Ma, Tong, and Xie]{rae}
Boyang Zheng, Nanye Ma, Shengbang Tong, and Saining Xie.
\newblock Diffusion transformers with representation autoencoders.
\newblock In \emph{International Conference on Learning Representations}, 2026.

\bibitem[Zheng et~al.()Zheng, Nie, Vahdat, and Anandkumar]{maskdit}
Hongkai Zheng, Weili Nie, Arash Vahdat, and Anima Anandkumar.
\newblock Fast training of diffusion models with masked transformers.
\newblock In \emph{arXiv preprint arXiv:2306.09305}.

\bibitem[Zhu et~al.()Zhu, Pan, Li, Yao, Sun, Mei, and Chen]{sd-dit}
Rui Zhu, Yingwei Pan, Yehao Li, Ting Yao, Zhenglong Sun, Tao Mei, and Chang~Wen Chen.
\newblock Sd-dit: Unleashing the power of self-supervised discrimination in diffusion transformer.
\newblock In \emph{CVPR}.

\end{thebibliography}
\bibliographystyle{iclr2027_conference}

\clearpage

\appendix

\section{Experimental setup}
\label{app:training_evaluation}
\textbf{Hyperparameter setup.}
Table~\ref{tab:cova_hyper} summarizes the configurations used by CoVA across
the three SiT model scales. All models operate on pre-extracted
$4\times32\times32$ VAE latents and use the same $2\times2$ latent
patchification. The composer is kept fixed across scales, while the SiT
alignment layer follows the layer choices used in the main experiments.
Unless otherwise specified, all models are trained for $400$K iterations with
AdamW~\cite{adam2}, a learning rate of $10^{-4}$, zero weight decay, and a
global batch size of $256$. The alignment loss is active for the first
$200$K iterations. SLC uses a constant replacement probability of $0.9$
and is activated only during the final $50$K iterations. Specifically, SLC
starts at $350$K iterations for the default $400$K-run setting, and at
$T-50$K for longer runs with a total training budget of $T$ iterations.
SLC uses the magnitude correction described in Sec.~\ref{sec:stage}.
Unless explicitly marked as \emph{naive SLC}, all CoVA results use the
magnitude-corrected soft condition in Eq.~\ref{eq:soft_label_condition}. Naive SLC is used only in diagnostic analyses, including the scheduling study in Fig.\ref{fig:ablation}(c) and the direction–magnitude analysis in Sec. ~\ref{sec:slc_analysis}.

\begin{table}[h]
    \centering
    \scriptsize
    \setlength{\tabcolsep}{3pt}
    \renewcommand{\arraystretch}{0.9}
    \caption{\textbf{CoVA hyperparameter settings across SiT model scales.}
    The alignment layer is indexed by the one-based SiT Transformer block
    depth. The same composer and stage-wise schedule are used across scales
    unless otherwise specified.}
    \resizebox{0.8\textwidth}{!}{%
    \begin{tabular}{lccc}
        \toprule
        \textbf{Backbone} & \textbf{SiT-B/2} & \textbf{SiT-L/2} & \textbf{SiT-XL/2} \\
        \midrule
        \textbf{Architecture} & & & \\
        \# Params & 130M & 458M & 675M \\
        Input latent & $32\times32\times4$ & $32\times32\times4$ & $32\times32\times4$ \\
        \# SiT blocks & 12 & 24 & 28 \\
        Hidden dimension & 768 & 1,024 & 1,152 \\
        \# Attention heads & 12 & 16 & 16 \\
        Latent patch size & $2\times2$ & $2\times2$ & $2\times2$ \\
        \midrule
        \textbf{Composer} & & & \\
        Feature dimension & 192 & 192 & 192 \\
        \# Composer blocks & 12 & 12 & 12 \\
        \# Composer heads & 3 & 3 & 3 \\
        MLP ratio & 4.0 & 4.0 & 4.0 \\
        Label smoothing & 0.1 & 0.1 & 0.1 \\
        Masking ratio & 0.7 & 0.7 & 0.7 \\
        $\lambda_{\mathrm{mim}}$ & 1.0 & 1.0 & 1.0 \\
        $\lambda_{\mathrm{flip}}$ & 1.0 & 1.0 & 1.0 \\
        \midrule
        \textbf{Alignment and conditioning} & & & \\
        Alignment layer & 4 & 8 & 8 \\
        Alignment weight & 1.0 & 1.0 & 1.0 \\
        Alignment duration & 200K & 200K & 200K \\
        SLC replacement probability & 0.9 & 0.9 & 0.9 \\
        SLC warmup & 350K & 350K & 350K \\
        \midrule
        \textbf{Optimization and sampling} & & & \\
        Global batch size & 256 & 256 & 256 \\
        Optimizer & AdamW & AdamW & AdamW \\
        Learning rate & $1\mathrm{e}{-4}$ & $1\mathrm{e}{-4}$ & $1\mathrm{e}{-4}$ \\
        Prediction & velocity & velocity & velocity \\
        Sampler & Euler-Maruyama & Euler-Maruyama & Euler-Maruyama \\
        Sampling steps & 250 & 250 & 250 \\
        \bottomrule
    \end{tabular}
    }
    \label{tab:cova_hyper}
\end{table}

\paragraph{Masked latent reconstruction.}
To encourage the latent composer to preserve local and spatial information from the VAE representation, we adopt a masked latent reconstruction objective. We randomly mask a subset of latent tokens, denoted by $\mathcal{M}$, and use the composer to infer their representations from the remaining visible tokens. A lightweight prediction head $D(\cdot)$ maps the composed feature $h_i$ at each masked position back to the VAE latent space. The reconstruction loss is defined as
\begin{equation}
\mathcal{L}_{\mathrm{mim}}
=
\frac{1}{|\mathcal{M}|}
\sum_{i\in\mathcal{M}}
\left\|
D(h_i)-Z_i
\right\|_2^2 ,
\end{equation}
where $Z_i$ denotes the original VAE latent feature at position $i$. The loss is evaluated only over masked positions, encouraging the composer to capture contextual and spatial structure rather than relying solely on local appearance.

\paragraph{Horizontal-flip consistency.}
We further impose a flip-consistency objective to encourage the composed dense representation to be equivariant to horizontal flipping. Given an input latent representation and its horizontally flipped counterpart, we obtain the corresponding composer features $h$ and $h^{\mathrm{flip}}$. We then spatially flip $h^{\mathrm{flip}}$ back to the original coordinate system using $\mathcal{F}^{-1}(\cdot)$ and minimize their cosine distance:
\begin{equation}
\mathcal{L}_{\mathrm{flip}}
=
\frac{1}{N}
\sum_i
d_{\cos}
\left(
h_i,\,
\mathcal{F}^{-1}(h_i^{\mathrm{flip}})
\right).
\end{equation}
Here, $N$ denotes the number of spatial tokens and $d_{\cos}(\cdot,\cdot)$ denotes cosine distance. This objective encourages the composer to retain spatially consistent and transformation-equivariant information in its dense representations.

\section{Spatial LDS Protocol}
\label{app:spatial_lds}
\paragraph{Metric and evaluation protocol.}
We measure the spatial organization of composer features using Spatial Local
Distance Similarity (Spatial LDS)~\cite{irepa}. For each ImageNet validation
image, we extract a row-major $16\times16$ feature grid from the deterministic
mode of the pretrained VAE posterior. Each token is $\ell_2$-normalized. LDS
is the mean cosine similarity over unique horizontal/vertical 4-neighbor
pairs (Manhattan distance $1$, $480$ pairs) minus the mean over directed token
pairs with Manhattan distance at least $6$ ($53{,}300$ pairs), self-pairs are
excluded. Thus, LDS measures local-versus-distant similarity contrast rather
than overall feature similarity.

\begin{figure}[t]
    \centering
    \includegraphics[width=0.9\linewidth]{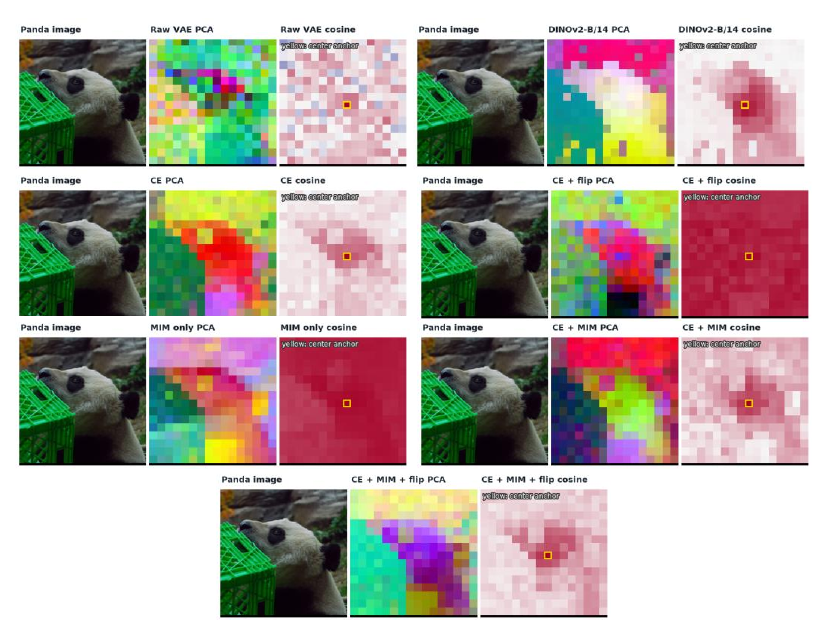}
    \caption{\textbf{PCA and cosine-map comparison for the seven table
    conditions.} Each panel uses the same panda image, $16\times16$ token
    grid, and center anchor.}
    \label{fig:ICLR4}
    \vspace{-10pt}
\end{figure}

\noindent We use the first $5{,}000$ ImageNet validation images, resized
and center-cropped to $256\times256$, identically across conditions. Composer
features are the unmasked dense tokens before the classification head.
DINOv2-B/14 is an external feature reference, with ImageNet normalization and
resizing to $224\times224$ to produce the same token grid. We evaluate the
200K model weights and report per-image means. The numerical comparison is summarized in
Table~\ref{tab:spatial_lds}, and the corresponding qualitative comparison is
shown in Figure~\ref{fig:ICLR4}.

\begin{table}[h]
    \centering
    \captionsetup{font={footnotesize}}
    \setlength{\tabcolsep}{4pt}
    \caption{\textbf{Spatial LDS objective comparisons at 200K.}
    Near/far denote mean cosine similarity; LDS is near minus far.
    All values are rounded to two decimal places.}
    \vspace{-3pt}
    \scriptsize
    \resizebox{0.45\linewidth}{!}{%
    \begin{tabular}{lccc}
        \toprule
        Feature / objective & Near$\uparrow$ & Far$\downarrow$ & LDS$\uparrow$ \\
        \midrule
        Raw VAE & 0.44 & 0.16 & 0.28 \\
        CE & 0.48 & 0.19 & 0.29 \\
        MIM only & 0.92 & 0.86 & 0.06 \\
        CE+MIM & 0.60 & 0.24 & 0.36 \\
        CE+flip & 0.92 & 0.83 & 0.09 \\
        CE+MIM+flip & 0.66 & 0.25 & \textbf{0.41} \\
        DINOv2-B/14 & 0.64 & 0.25 & 0.40 \\
        \bottomrule
    \end{tabular}}
    \label{tab:spatial_lds}
\end{table}

\noindent\textbf{The objective effects interact.}
Using CE as the anchor, adding MIM increases LDS by $+0.07$. Adding flip
consistency without MIM changes LDS by $-0.20$. When MIM is already present,
adding flip increases LDS by $+0.05$; equivalently, adding MIM to CE+flip
increases LDS by $+0.31$. The difference-in-differences interaction is
$+0.25$, indicating strong complementarity between MIM and flip consistency
in this run-level comparison.

\noindent LDS measures local-versus-distant contrast rather than overall
feature similarity. MIM-only and CE+flip have high near and far cosine
similarities ($0.92/0.86$ and $0.92/0.83$, respectively), but low LDS. Their
features are therefore highly similar across the grid and show limited
local-versus-distant differentiation. MIM supplies patch-level reconstruction
pressure, CE supplies class semantics, and flip consistency constrains how
dense features transform under reflection. The results suggest that flip
consistency is useful in this comparison when combined with MIM, but not in
isolation.

\section{Additional Ablation Studies}
\label{app:additional_ablations}
\noindent
All experiments in this section use SiT-B/2 trained for 400K iterations with 250 SDE sampling steps and without CFG, unless stated otherwise. The main paper presents the primary ablations on composer objectives, alignment duration, sample-adaptive versus fixed conditioning, and SLC warmup. Here, we provide additional analyses of composer capacity and robustness.

\noindent
{\setlength{\fboxsep}{1pt}%
\colorbox{yellow!12}{\textbf{Cumulative ablation of CoVA.}}}%
\hspace{0.2em}%
We progressively add the main components of CoVA to the SiT-B/2 baseline under
the same 400K-step training and evaluation protocol.
As shown in Table~\ref{tab:app_cova_cumulative}, composed representation
alignment first reduces FID from $33.02$ to $26.83$.
Early stopping the alignment further improves FID to $24.87$.
Finally, reusing the detached composer posterior for late-stage soft label
conditioning yields the full CoVA model and further reduces FID to $21.45$.
These results show that \textbf{representation composition, stage-adaptive
alignment, and late-stage conditioning provide complementary improvements and
jointly contribute to the final performance of CoVA.}

\begin{table}[t]
\centering
\caption{
\textbf{Cumulative ablation of CoVA on SiT-B/2.}
All models are trained for 400K iterations on ImageNet $256\times256$ without
CFG under the same optimization and sampling protocol.
Each row progressively adds one component to the preceding configuration.
}
\label{tab:app_cova_cumulative}
\setlength{\tabcolsep}{5pt}
\begin{tabular}{lccccc}
\toprule
Configuration
& FID$\downarrow$
& sFID$\downarrow$
& Prec.$\uparrow$
& Rec.$\uparrow$
& IS$\uparrow$ \\
\midrule
SiT-B/2
& 33.02 & 6.46 & 0.53 & 0.63 & 43.71 \\

+ Composed representation alignment
& 26.83 & 6.32 & 0.57 & 0.64 & 57.69 \\

+ Alignment early stopping
& 24.87 & 5.96 & 0.58 & 0.64 & 61.19 \\

+ Soft label conditioning
& 21.45 & 5.69 & 0.59 & 0.63 & 74.57 \\
\bottomrule
\end{tabular}
\end{table}


\noindent
{\setlength{\fboxsep}{1pt}%
\colorbox{blue!12}{\textbf{Composer capacity.}}}%
\hspace{0.2em}%
We first vary the composer capacity while keeping the training recipe fixed.
As shown in Table~\ref{tab:app_composer_capacity}, performance remains largely
stable across a wide range of composer sizes. Increasing the capacity from the
smallest to the largest configuration changes FID only marginally, from
$27.52$ to $27.32$, while the intermediate configuration obtains $28.19$.
The absence of a monotonic scaling trend suggests that \textbf{the gains of CoVA do
not rely on increasing auxiliary model capacity. A compact composer already
provides effective supervision.}

\begin{table}[H]
    \centering
    \captionsetup{font={footnotesize}}
    \setlength{\tabcolsep}{2.5pt}
    \caption{\textbf{Composer capacity.}
Configurations are reported as $(d_h,\mathrm{depth},\mathrm{heads},r_{\mathrm{MLP}})$.
Params, GFLOPs, and GMACs refer to the auxiliary composer.}
    \vspace{-3pt}
    \scriptsize
    \resizebox{0.75\linewidth}{!}{%
    \begin{tabular}{lccc|ccccc}
        \toprule
        Composer & Params & GFLOPs & GMACs
        & FID$\downarrow$ & sFID$\downarrow$
        & Prec.$\uparrow$ & Rec.$\uparrow$ & IS$\uparrow$ \\
        \midrule
        $(192,12,3,4)$ & 5.59M  & 3.34  & 1.67 & 27.66 & 6.32 & 0.57 & 0.64 & 55.28 \\
$(192, 8,3,4)$ & 3.81M  & 2.23  & 1.11 & 27.52 & 6.22 & 0.57 & 0.64 & 55.96 \\
$(768, 3,12,4)$ & 22.26M & 11.54 & 5.77 & 28.19 & 6.25 & 0.56 & 0.64 & 54.84 \\
$(768, 4,12,4)$ & 29.35M & 15.38 & 7.69 & 27.32 & 6.27 & 0.57 & 0.64 & 55.74 \\
        \bottomrule
    \end{tabular}
    }
    \label{tab:app_composer_capacity}
    \vspace{-0.8em}
\end{table}

\begin{table}[h]
    \centering
    \captionsetup{font={footnotesize}}
    \setlength{\tabcolsep}{4pt}
    \caption{\textbf{Sensitivity to composer objective hyperparameters.}
    We vary label smoothing $\eta$ and MIM masking ratio $\rho$ while keeping
    the remaining training settings fixed.}
    \vspace{-3pt}
    \scriptsize
    \resizebox{0.5\linewidth}{!}{%
    \begin{tabular}{c|ccccc}
        \toprule
        Setting
        & FID$\downarrow$ & sFID$\downarrow$
        & Prec.$\uparrow$ & Rec.$\uparrow$ & IS$\uparrow$ \\
        \midrule
        \multicolumn{6}{c}{\textit{Label smoothing}} \\
        \midrule
        $\eta=0.00$ & 31.91 & 6.78 & 0.54 & 0.64 & 48.64 \\
        $\eta=\mathbf{0.10}$ & 30.86 & 6.83 & 0.55 & 0.64 & 50.95 \\
        $\eta=0.20$ & \textbf{30.76} & 6.84 & 0.55 & 0.63 & 50.45 \\
        $\eta=0.30$ & 31.36 & 6.79 & 0.54 & 0.64 & 49.90 \\
        \midrule
        \multicolumn{6}{c}{\textit{MIM masking ratio}} \\
        \midrule
        $\rho=0.10$ & 31.63 & 6.74 & 0.54 & 0.64 & 49.67 \\
        $\rho=0.30$ & 30.01 & 6.82 & 0.55 & 0.64 & 51.91 \\
        $\rho=0.50$ & 28.56 & 6.53 & 0.56 & 0.64 & 53.96 \\
        $\rho=\mathbf{0.70}$ & \textbf{28.11} & \textbf{6.41}
        & \textbf{0.56} & \textbf{0.64} & \textbf{54.92} \\
        \bottomrule
    \end{tabular}}
    \label{tab:app_composer_hyperparameters}
\end{table}

\noindent
{\setlength{\fboxsep}{1pt}%
\colorbox{orange!18}{\textbf{Sensitivity to composer objective hyperparameters.}}}%
\hspace{0.2em}%
We further examine the sensitivity of CoVA to the main composer-objective
hyperparameters. As shown in Table~\ref{tab:app_composer_hyperparameters}, for label smoothing, moderate values are consistently better
than either no smoothing or stronger smoothing: $\eta=0.10$ and $0.20$ achieve
similar performance, reducing FID from $31.91$ to $30.86$ and $30.76$,
respectively, while $\eta=0.30$ slightly degrades performance. The MIM masking
ratio has a clearer effect, with FID steadily improving from $31.63$ at
$\rho=0.10$ to $28.11$ at $\rho=0.70$. Overall, CoVA is relatively insensitive
to moderate changes in label smoothing, while benefiting from stronger masked
reconstruction. We use $\eta=0.10$ and $\rho=0.70$ as the default settings.

{\setlength{\fboxsep}{1pt}%
\colorbox{green!12}{\textbf{Alignment placement and loss weights.}}}%
\hspace{0.2em}%
Table~\ref{tab:app_alignment_settings} examines two implementation choices independently. Moving the alignment target from block $4$ to deeper blocks consistently degrades performance, with FID increasing from $27.66$ to $30.02$ and $32.65$ at blocks $6$ and $8$, respectively. In contrast, moderate changes to the alignment, MIM, and flip-consistency weights have a smaller effect, although none improves upon the default configuration. These results indicate that \textbf{CoVA does not require careful tuning of the auxiliary loss coefficients.}

\begin{table}[h]
    \centering
    \captionsetup{font={footnotesize}}
    \setlength{\tabcolsep}{3.5pt}
    \caption{\textbf{Sensitivity to alignment placement and loss weights.}
    Each sweep varies one factor while keeping the remaining settings fixed.
    The default alignment block is $4$, and all default loss weights are $1.0$.}
    \vspace{-3pt}
    \scriptsize
    \resizebox{0.5\linewidth}{!}{%
    \begin{tabular}{lccccc}
        \toprule
        Setting
        & FID$\downarrow$ & sFID$\downarrow$
        & Prec.$\uparrow$ & Rec.$\uparrow$ & IS$\uparrow$ \\
        \midrule

        \multicolumn{6}{c}{\textit{(a) Alignment block}} \\
        \midrule
        Block $4$ (Default)
        & 27.66 & 6.32 & 0.57 & 0.64 & 55.28 \\
        Block $6$
        & 30.02 & 6.43 & 0.55 & 0.64 & 51.09 \\
        Block $8$
        & 32.65 & 6.42 & 0.54 & 0.64 & 47.13 \\

        \midrule
        \multicolumn{6}{c}{\textit{(b) Loss weights}} \\
        \midrule
        Default $(1,1,1)$
        & 27.66 & 6.32 & 0.57 & 0.64 & 55.28 \\
        $\lambda_{\rm align}=0.5$
        & 28.25 & 6.39 & 0.56 & 0.64 & 54.45 \\
        $\lambda_{\rm mim}=5$
        & 28.73 & 6.39 & 0.56 & 0.64 & 53.75 \\
        $\lambda_{\rm flip}=5$
        & 28.69 & 6.44 & 0.56 & 0.63 & 53.33 \\
        \bottomrule
    \end{tabular}}
    \label{tab:app_alignment_settings}
\end{table}

{\setlength{\fboxsep}{1pt}%
\colorbox{violet!12}{\textbf{Online adaptation and SLC scheduling.}}}%
\hspace{0.2em}%
Table~\ref{tab:app_conditioning_schedules} examines whether the composer should
remain online and whether SLC requires more elaborate scheduling. Freezing a
pre-trained composer degrades FID from 27.66 to 29.27, supporting joint online
adaptation throughout training. For SLC, early or middle activation performs
poorly, while delaying it to the later stage is substantially better.
Conditioning the replacement probability on flow time $t$ does not outperform
the simple late-stage schedule, and Dirichlet conditioning is considerably
worse. These results support a simple design: keep the composer adaptive and
introduce SLC only at a late training stage, without additional
$t$-dependent conditioning schedules.

\begin{table}[h]
    \centering
    \captionsetup{font={footnotesize}}
    \setlength{\tabcolsep}{3.5pt}
    \caption{\textbf{Online adaptation and alternative SLC schedules.}
    These schedules are exploratory variants of the final late-stage SLC design.}
    \vspace{-3pt}
    \scriptsize
    \resizebox{0.55\linewidth}{!}{%
    \begin{tabular}{lccccc}
        \toprule
        Setting
        & FID$\downarrow$ & sFID$\downarrow$
        & Prec.$\uparrow$ & Rec.$\uparrow$ & IS$\uparrow$ \\
        \midrule
        \multicolumn{6}{c}{\textit{Composer adaptation}} \\
        \midrule
        Joint online composer
        & 27.66 & 6.32 & 0.57 & 0.64 & 55.28 \\
        Pre-train then freeze
        & 29.27 & 6.40 & 0.56 & 0.63 & 52.30 \\
        \midrule
        \multicolumn{6}{c}{\textit{Alternative SLC schedules}} \\
        \midrule
        Early activation ($150$K)
        & 35.57 & 6.49 & 0.52 & 0.64 & 43.31 \\
        Middle activation
        & 35.96 & 6.48 & 0.52 & 0.64 & 43.02 \\
        Late activation
        & 30.57 & 6.14 & 0.55 & 0.63 & 53.25 \\
        $p\propto t$
        & 31.97 & 6.38 & 0.54 & 0.63 & 49.80 \\
        $p\propto(1-t)$
        & 29.23 & 6.01 & 0.56 & 0.62 & 54.88 \\
        Dirichlet, $p=0.30$
        & 38.80 & 6.40 & 0.51 & 0.63 & 36.61 \\
        \bottomrule
    \end{tabular}}
    \label{tab:app_conditioning_schedules}
\end{table}

{\setlength{\fboxsep}{1pt}%
\colorbox{yellow!20}{\textbf{CFG sensitivity after SLC training.}}}%
\hspace{0.2em}%
We evaluate the robustness of the SLC-trained SiT-XL/2 model to
inference-time CFG settings. As shown in
Table~\ref{tab:app_cfg_sweep}, performance is stable over a moderate range of
guidance scales. With the guidance interval fixed to $[0,0.50]$, scales from
$1.60$ to $1.80$ achieve FIDs between $1.99$ and $2.04$, indicating little
sensitivity to precise CFG tuning. The lowest FID of $1.99$ is obtained at
scale $1.65$, while scale $1.80$ yields substantially higher IS and slightly
higher Precision with only a minor FID difference. We therefore use
$\mathrm{CFG}=1.80$ with interval $[0,0.50]$ for the reported results.

\begin{table}[h]
    \centering
    \captionsetup{font={footnotesize}}
    \setlength{\tabcolsep}{4pt}
    \caption{\textbf{CFG sensitivity of SLC-trained SiT-XL/2.}
    Results are evaluated after $2$M training iterations using $50$K samples.
    The interval denotes the portion of the sampling trajectory where CFG is applied.}
    \vspace{-3pt}
    \scriptsize
    \resizebox{0.55\linewidth}{!}{%
    \begin{tabular}{cc|ccccc}
        \toprule
        CFG scale & Interval
        & FID$\downarrow$ & sFID$\downarrow$
        & Prec.$\uparrow$ & Rec.$\uparrow$ & IS$\uparrow$ \\
        \midrule
        \multicolumn{7}{c}{\textit{Guidance-scale sensitivity}} \\
        \midrule
        1.60 & $[0,0.50]$
        & 2.01 & 4.63 & 0.79 & \textbf{0.62} & 291.74 \\
        1.65 & $[0,0.50]$
        & \textbf{1.99} & 4.54 & 0.79 & 0.61 & 299.15 \\
        1.70 & $[0,0.50]$
        & 2.01 & 4.56 & 0.79 & \textbf{0.62} & 304.19 \\
        1.80 & $[0,0.50]$
        & 2.03 & \textbf{4.52} & \textbf{0.80} & 0.62 & \textbf{316.42} \\
        \midrule
        \multicolumn{7}{c}{\textit{Guidance-interval sensitivity}} \\
        \midrule
        1.80 & $[0,0.40]$
        & 2.12 & 4.61 & 0.78 & \textbf{0.62} & 278.18 \\
        1.80 & $[0,0.50]$
        & 2.03 & \textbf{4.52} & \textbf{0.80} & \textbf{0.62} & \textbf{316.42} \\
        \bottomrule
    \end{tabular}}
    \label{tab:app_cfg_sweep}
\end{table}

\section{Understanding the Magnitude Confound in Soft Label Conditioning}
\label{app:slc_magnitude_analysis}

{\setlength{\fboxsep}{1pt}%
\colorbox{violet!12}{\textbf{Geometry of soft class embeddings.}}}%
\hspace{0.2em}%
We first analyze why posterior averaging systematically changes the magnitude
of the class condition. Let $\{\boldsymbol e_k\}_{k=1}^{K}$ denote the learned
class embeddings and let
\begin{equation}
    \boldsymbol e_s
    =
    \sum_{k=1}^{K} q_k \boldsymbol e_k
\end{equation}
be the naive soft condition produced from the composer posterior
$\boldsymbol q$.
For the trained SiT-XL/2 model, the class embeddings have similar magnitudes,
with a mean L2 norm of $1.542$ and a standard deviation of $0.180$.
Meanwhile, different class embeddings are nearly orthogonal: the mean
pairwise cosine similarity is only $0.016\pm0.078$.

The squared magnitude of the soft condition can be written as
\begin{equation}
\begin{aligned}
    \|\boldsymbol e_s\|_2^2
    &=
    \left\|
    \sum_k q_k \boldsymbol e_k
    \right\|_2^2 \\
    &=
    \sum_k q_k^2\|\boldsymbol e_k\|_2^2
    +
    \sum_{i\neq j}
    q_iq_j
    \boldsymbol e_i^\top\boldsymbol e_j .
\end{aligned}
\label{eq:soft_norm_exact}
\end{equation}
Since the class embeddings have comparable norms and small pairwise inner
products, the cross terms in Eq.~\ref{eq:soft_norm_exact} are relatively
small. Denoting the typical class-embedding norm by $\bar r$, we obtain the
approximation
\begin{equation}
    \|\boldsymbol e_s\|_2
    \approx
    \bar r
    \sqrt{\sum_k q_k^2}
    =
    \bar r \|\boldsymbol q\|_2 .
\label{eq:soft_norm_approx}
\end{equation}
Therefore, posterior averaging naturally couples semantic uncertainty with
conditioning magnitude. A concentrated posterior approaches the magnitude of
a hard-label embedding, whereas a more dispersed posterior produces a shorter
conditioning vector. Consequently, naive SLC changes not only the semantic
direction of the class condition, but also its effective magnitude.

{\setlength{\fboxsep}{1pt}%
\colorbox{violet!12}{\textbf{Connection to classifier-free guidance.}}}%
\hspace{0.2em}%
The sensitivity to condition magnitude can also be related to classifier-free
guidance (CFG). Let
$\boldsymbol v(\boldsymbol x,t,\boldsymbol e)$ denote the velocity prediction
under condition $\boldsymbol e$, and let $\boldsymbol e_{\mathrm{null}}$
denote the unconditional embedding. CFG produces
\begin{equation}
    \boldsymbol v_{\mathrm{CFG}}
    =
    \boldsymbol v(\boldsymbol x,t,\boldsymbol e_{\mathrm{null}})
    +
    w
    \left[
    \boldsymbol v(\boldsymbol x,t,\boldsymbol e)
    -
    \boldsymbol v(\boldsymbol x,t,\boldsymbol e_{\mathrm{null}})
    \right],
\label{eq:cfg_output}
\end{equation}
where $w$ is the guidance scale.

Locally linearizing the conditional prediction around
$\boldsymbol e_{\mathrm{null}}$ gives
\begin{equation}
    \boldsymbol v(\boldsymbol x,t,\boldsymbol e)
    \approx
    \boldsymbol v(\boldsymbol x,t,\boldsymbol e_{\mathrm{null}})
    +
    \boldsymbol J_e
    \left(
    \boldsymbol e-\boldsymbol e_{\mathrm{null}}
    \right),
\end{equation}
where
$\boldsymbol J_e =
\partial \boldsymbol v / \partial \boldsymbol e$.
Substituting this approximation into Eq.~\ref{eq:cfg_output} yields
\begin{equation}
\begin{aligned}
    \boldsymbol v_{\mathrm{CFG}}
    &\approx
    \boldsymbol v(\boldsymbol x,t,\boldsymbol e_{\mathrm{null}})
    +
    w\boldsymbol J_e
    \left(
    \boldsymbol e-\boldsymbol e_{\mathrm{null}}
    \right) \\
    &\approx
    \boldsymbol v
    \left(
    \boldsymbol x,
    t,
    \boldsymbol e_{\mathrm{null}}
    +
    w
    \left(
    \boldsymbol e-\boldsymbol e_{\mathrm{null}}
    \right)
    \right).
\end{aligned}
\label{eq:cfg_embedding_view}
\end{equation}
This provides a useful local interpretation of CFG as scaling the displacement
of the condition from the unconditional anchor in embedding space. It does
\emph{not}, however, imply an exact equivalence between condition magnitude
and CFG. The approximation is local, the soft condition is not generally
collinear with the hard condition, and the limiting average class embedding
does not coincide with the learned unconditional embedding. Moreover, the
conditional response of a deep network is nonlinear, so
$\boldsymbol J_e$ need not remain constant for large changes in condition
magnitude.

Nevertheless, Eq.~\ref{eq:cfg_embedding_view} suggests that condition
magnitude can act as an effective conditioning-strength coordinate. This is
consistent with our inference-time scaling experiment, where directly
rescaling a hard-label embedding substantially changes the
fidelity--coverage trade-off even though all model parameters are fixed.

{\setlength{\fboxsep}{1pt}%
\colorbox{violet!12}{\textbf{Magnitude and direction play distinct roles.}}}%
\hspace{0.2em}%
The analysis above explains why naive SLC contains an implicit
conditioning-strength component, but does not imply that its directional
variation is irrelevant. Our direction--magnitude decomposition in the main
paper shows that replacing the hard-label magnitude with the attenuated soft
magnitude produces a large improvement in FID, but also noticeably reduces
Recall. This behavior indicates that magnitude attenuation primarily acts as
a fidelity-oriented conditioning shortcut. In contrast, preserving the soft
direction while restoring the hard-label magnitude yields a smaller but
meaningful improvement while retaining substantially better coverage.
Furthermore, shuffling soft directions across samples strongly degrades
performance, demonstrating that the posterior-dependent direction contains
sample-specific semantic information.

Taken together, naive SLC entangles two fundamentally different effects:
\emph{posterior-dependent semantic direction} and
\emph{posterior-dependent conditioning magnitude}. The latter can strongly
alter the fidelity--coverage balance and therefore confounds the apparent gain
of soft conditioning. CoVA explicitly removes this confound by restoring the
soft condition to the corresponding hard-label magnitude, while preserving
its posterior-dependent direction. In this way, the soft condition contributes
sample-specific semantic information without relying on unintended changes in
conditioning strength.

\noindent
{\setlength{\fboxsep}{1pt}%
\colorbox{blue!12}{\textbf{Null-centered controls for conditioning strength.}}}%
\hspace{0.2em}%
Our default magnitude correction in Eq.~(8) restores the soft condition to the
norm of the corresponding hard-label embedding. Since the local interpretation
of classifier-free guidance is more naturally expressed relative to the learned
unconditional embedding $\boldsymbol{e}_{\varnothing}$, we further examine
whether the observed effect depends on measuring magnitude from the origin.
We conduct two complementary null-centered controls, one during training and
one only at inference.

\paragraph{Null-centered normalization during training.}
For the naive soft condition
$\bar{\boldsymbol{c}}_j=\boldsymbol{q}_j^{\top}\boldsymbol{E}$,
we construct a null-centered variant that explicitly preserves the distance
from the learned unconditional embedding:
\begin{equation}
\boldsymbol{c}^{\mathrm{soft,null}}_j
=
\boldsymbol{e}_{\varnothing}
+
\frac{
\left\|
\boldsymbol{E}_{y_j}-\boldsymbol{e}_{\varnothing}
\right\|_2
}{
\left\|
\bar{\boldsymbol{c}}_j-\boldsymbol{e}_{\varnothing}
\right\|_2+\epsilon
}
\left(
\bar{\boldsymbol{c}}_j-\boldsymbol{e}_{\varnothing}
\right).
\label{eq:null_centered_slc}
\end{equation}
By construction,
\begin{equation}
\left\|
\boldsymbol{c}^{\mathrm{soft,null}}_j
-
\boldsymbol{e}_{\varnothing}
\right\|_2
=
\left\|
\boldsymbol{E}_{y_j}
-
\boldsymbol{e}_{\varnothing}
\right\|_2,
\end{equation}
so the soft condition preserves the hard-label displacement from the
unconditional anchor rather than only its norm with respect to the origin.

Starting from the same 350K checkpoint, we replace the default normalization
with Eq.~\ref{eq:null_centered_slc} during the final 50K training iterations,
while keeping all other SLC settings unchanged.
As shown in Table~\ref{tab:app_null_centered_controls}, null-centered
normalization achieves an FID of $21.14$, closely matching the $21.45$ FID of
the default CoVA formulation. It also retains high Recall ($0.628$).
This result indicates that \textbf{the improvement from corrected SLC does not
depend on using the embedding-space origin as the normalization reference};
preserving conditioning strength relative to the learned unconditional
embedding leads to essentially the same behavior.

\paragraph{Null-centered scaling at inference.}
We next directly perturb conditioning strength relative to the unconditional
anchor while keeping the model parameters fixed.
For a hard-label embedding $\boldsymbol{e}_y$, we define
\begin{equation}
\boldsymbol{e}_{y}^{(\gamma)}
=
\boldsymbol{e}_{\varnothing}
+
\gamma
\left(
\boldsymbol{e}_y-\boldsymbol{e}_{\varnothing}
\right),
\label{eq:null_centered_scaling}
\end{equation}
where $\gamma=1$ recovers the standard hard-label condition.
This transformation changes only the displacement from the unconditional
embedding and therefore provides a direct intervention on the
conditioning-strength coordinate suggested by the local CFG interpretation.

As $\gamma$ increases from $1.0$ to $1.35$, FID decreases from $24.87$ to
$9.19$ and Precision increases from $0.58$ to $0.71$, while Recall decreases
from $0.64$ to $0.56$.
The same monotonic fidelity--coverage trade-off observed with origin-centered
scaling therefore remains when conditioning strength is defined relative to
$\boldsymbol{e}_{\varnothing}$.
Together with the training-time control above, these results confirm that
\textbf{conditioning magnitude constitutes a genuine conditioning-strength
coordinate rather than an artifact of origin-centered normalization}, while
the remaining gain of magnitude-corrected SLC is attributable to the
posterior-dependent directional signal.

\begin{table}[t]
\centering
\caption{
\textbf{Null-centered controls for soft label conditioning.}
Top: inference-time scaling of the hard-label displacement from the learned
unconditional embedding using Eq.~\ref{eq:null_centered_scaling}; all model
parameters are fixed and the model is trained without SLC.
Bottom: training-time SLC using the null-centered magnitude correction in
Eq.~\ref{eq:null_centered_slc}, applied during the final 50K iterations.
}
\label{tab:app_null_centered_controls}
\setlength{\tabcolsep}{5pt}
\begin{tabular}{lccccc}
\toprule
Setting
& FID$\downarrow$
& sFID$\downarrow$
& Prec.$\uparrow$
& Rec.$\uparrow$
& IS$\uparrow$ \\
\midrule
\multicolumn{6}{l}{\textit{Inference-time null-centered scaling}} \\
$\gamma=1.00$
& 24.87 & 5.95 & 0.58 & 0.64 & 61.19 \\
$\gamma=1.10$
& 18.48 & 5.72 & 0.62 & 0.61 & 81.05 \\
$\gamma=1.20$
& 13.73 & 5.55 & 0.66 & 0.59 & 104.42 \\
$\gamma=1.30$
& 10.41 & 5.47 & 0.70 & 0.57 & 129.03 \\
$\gamma=1.35$
& \textbf{9.19} & \textbf{5.45} & \textbf{0.71} & 0.56 & \textbf{141.10} \\
\midrule
\multicolumn{6}{l}{\textit{Training-time null-centered SLC}} \\
Null-centered magnitude correction
& 21.14 & 5.64 & 0.59 & 0.63 & 75.19 \\
\bottomrule
\end{tabular}
\end{table}


\section{Supplementary Qualitative Results}
\label{app:qualitative_results}

\begin{figure}[ht]
    \centering
    \includegraphics[width=0.9\linewidth]{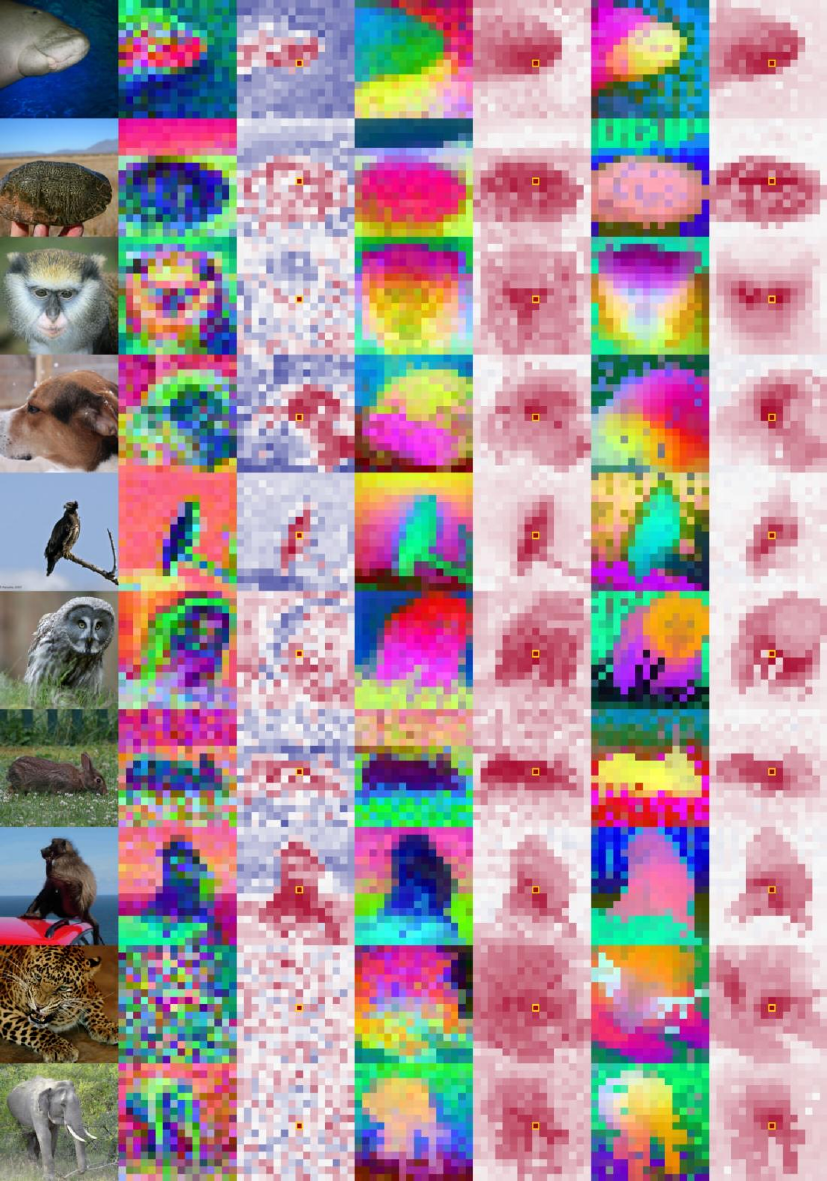}
    \vspace{-5pt}
    \caption{\textbf{Additional visualizations of spatial representations.} }
    \label{fig:PCA2}
    \vspace{-1em}
\end{figure}

We provide additional qualitative results for both image generation and
spatial representation quality. Figure~\ref{fig:abstract-fig}
shows
randomly sampled images generated by SiT-XL/2 with CoVA, while
Figure~\ref{fig:PCA2} extends the PCA and patch-level cosine-similarity
analysis in Figure~\ref{fig:PCA} to a broader set of ImageNet examples.

\begin{figure}[ht]
    \centering
    \includegraphics[width=0.9\linewidth]{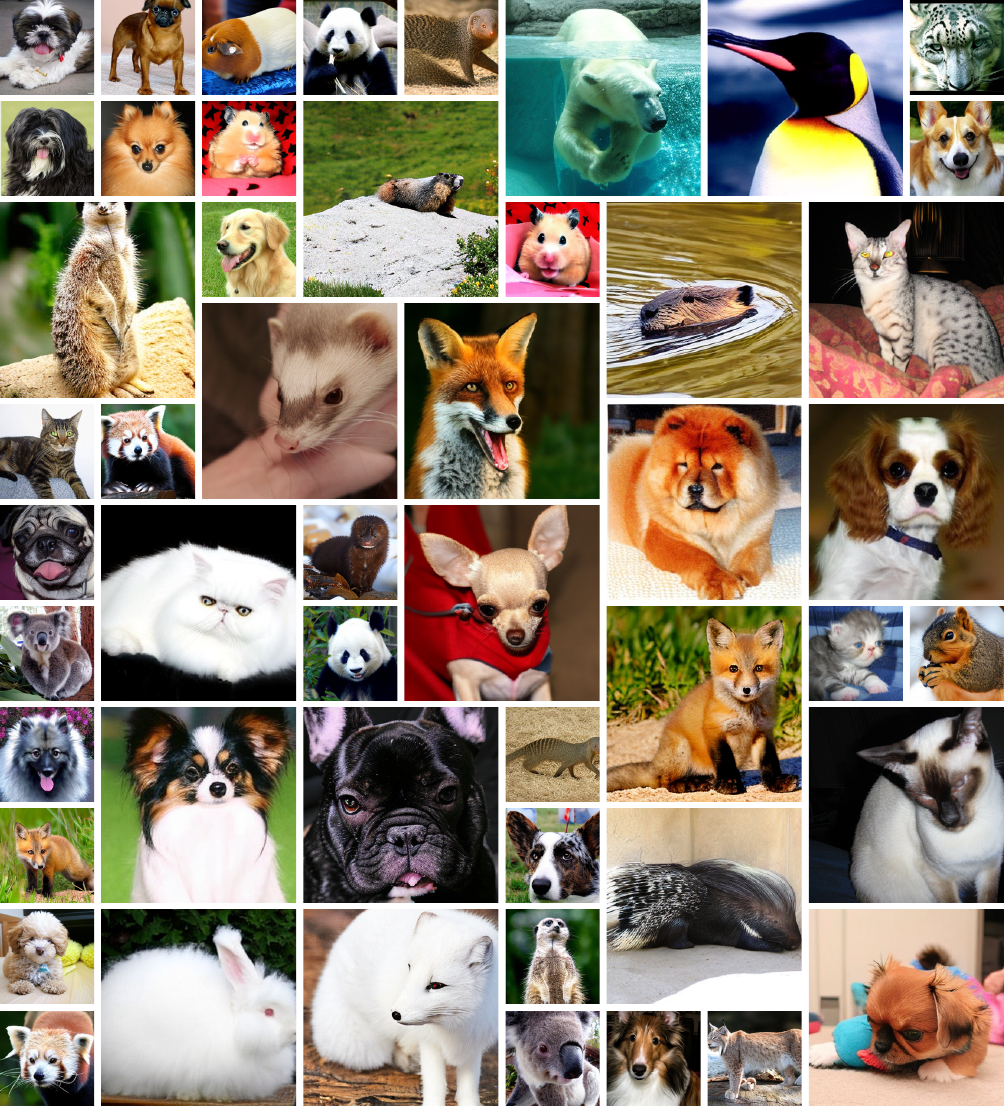}
    \vspace{-5pt}
    \caption{\textbf{The visualization results of SiT-XL/2 + CoVA use CFG with w = 4.0.} }
    \label{fig:abstract-fig}
    \vspace{-1em}
\end{figure}

\begin{figure}[ht]
    \centering
    \includegraphics[width=0.9\linewidth]{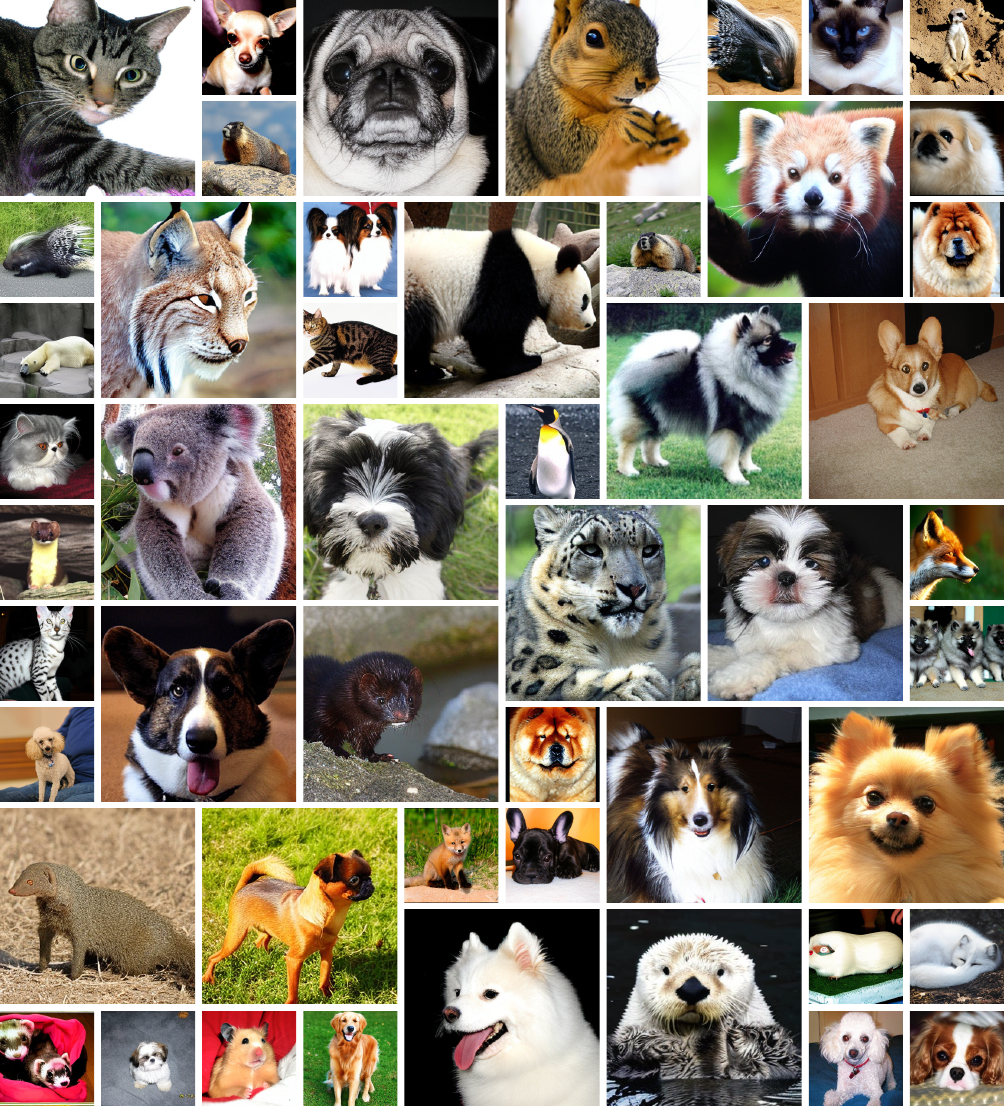}
    \vspace{-5pt}
    \caption{\textbf{The visualization results of SiT-XL/2 + CoVA use CFG with w = 4.0.} }
    \label{fig:abstract-fig}
    \vspace{-1em}
\end{figure}

\begin{figure}[ht]
    \centering
    \includegraphics[width=0.9\linewidth]{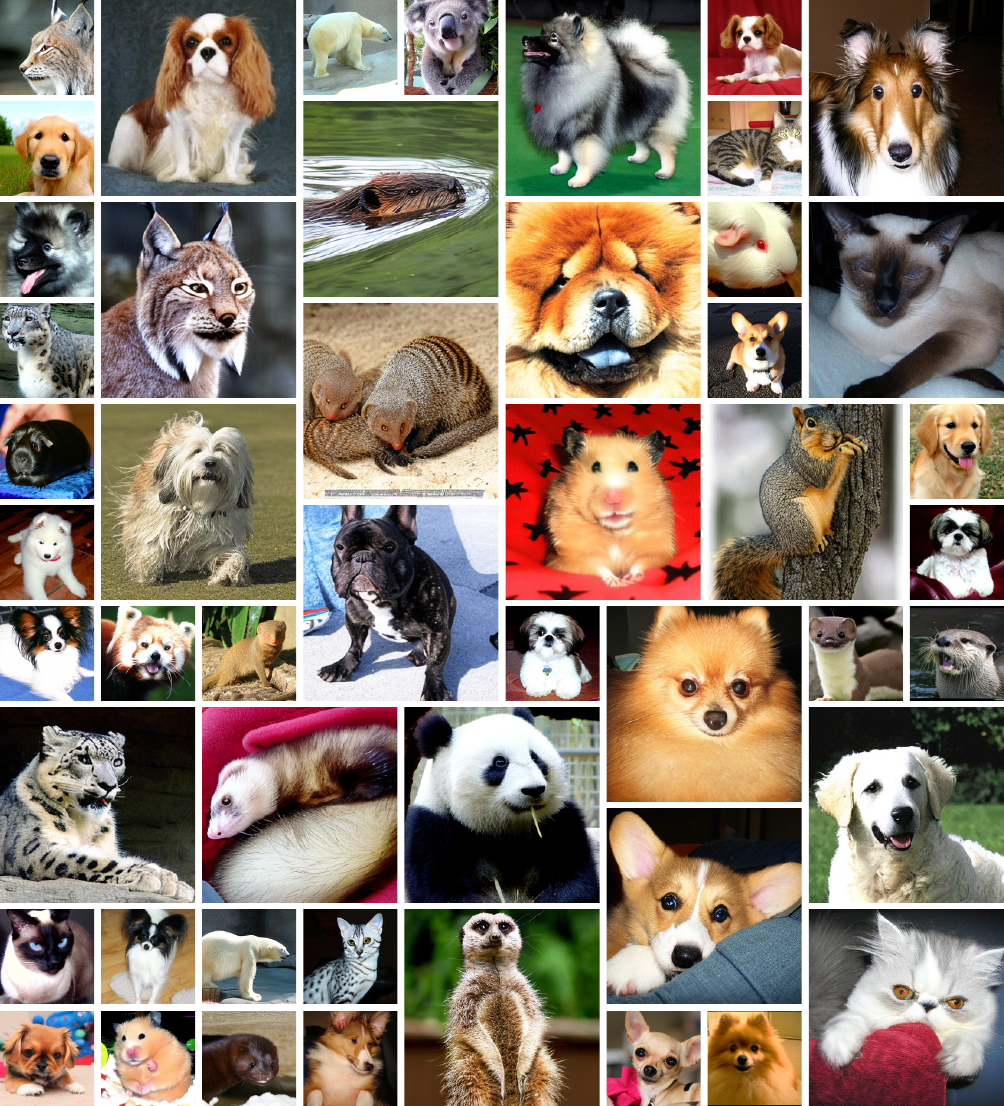}
    \vspace{-5pt}
    \caption{\textbf{The visualization results of SiT-XL/2 + CoVA use CFG with w = 4.0.} }
    \label{fig:abstract-fig}
    \vspace{-1em}
\end{figure}

\end{document}